\documentclass{article}
\usepackage[english]{babel}
\usepackage[T1]{fontenc}
\usepackage{authblk}
\usepackage{biblatex}
\usepackage{graphicx}
\usepackage{subcaption}
\usepackage{adjustbox}
\usepackage{multirow}
\usepackage{listings}
\usepackage{hyperref}
\usepackage{caption}
\usepackage{tikz,amsmath}
\usepackage{makecell}
\usepackage[newfloat,frozencache]{minted}
\newenvironment{code}{\captionsetup{type=listing}}{}
\SetupFloatingEnvironment{listing}{name=Listing}

\providecommand{\keywords}[1]
{
  \small	
  \textbf{\textit{Keywords---}} #1
}

\begin{document}
\title{DocXPand-25k: a large and diverse benchmark dataset for identity documents analysis}
\date{}
\author[1]{Julien Lerouge \footnote{corresponding author \href{mailto:julien.lerouge@quicksign.com}{julien.lerouge@quicksign.com}}}
\author[1]{Guillaume Betmont}
\author[1]{Thomas Bres}
\author[1]{Evgeny Stepankevich}
\author[1]{Alexis Bergès}
\affil[1]{QuickSign, 19 rue Poissonnière 75002 Paris}

\maketitle

\begin{abstract}
    Identity document (ID) image analysis has become essential for many online services, like bank account opening or insurance subscription. In recent years, much research has been conducted on subjects like document localization, text recognition and fraud detection, to achieve a level of accuracy reliable enough to automatize identity verification. However, there are only a few available datasets to benchmark ID analysis methods, mainly because of privacy restrictions, security requirements and legal reasons.
    In this paper, we present the DocXPand-25k dataset, which consists of 24\,994 richly labeled IDs images, generated using custom-made vectorial templates representing nine fictitious ID designs, including four identity cards, two residence permits and three passports designs. These synthetic IDs feature artificially generated personal information (names, dates, identifiers, faces, barcodes, \dots), and present a rich diversity in the visual layouts and textual contents.
    We collected about 5.8k diverse backgrounds coming from real-world photos, scans and screenshots of IDs to guarantee the variety of the backgrounds. The software we wrote to generate these images has been published\footnote{\href{https://github.com/QuickSign/docxpand/}{https://github.com/QuickSign/docxpand/}} under the terms of the MIT license, and our dataset has been published\footnote{\href{https://github.com/QuickSign/docxpand/releases/tag/v1.0.0}{https://github.com/QuickSign/docxpand/releases/tag/v1.0.0}} under the terms of the CC-BY-NC-SA 4.0 License.
\end{abstract}

\section{Introduction}

The banking and insurance actors have to comply to various laws and regulations, certify the authenticity of proof documents (identity documents, proof of residence documents, etc.) and the identity of their customers, following the "Know Your Customer" rules. Automatic document analysis methods, including document classification, localization, text recognition and visual elements extraction have been developed over the years to decrease the time and human efforts needed to process these documents.
Identity documents (ID) recognition systems~\cite{attivissimo2019automatic, chiron2021id} are now widely used. In the last few years, a significant amount of research has been published in the topic of IDs analysis on mobile terminals~\cite{bulatov2017smart, skoryukina2019fast, wu2022ldrnet}. However, despite numerous publications on the topic of IDs analysis, due to privacy restrictions, security requirements and legal reasons, researchers are constrained by the scarcity of large public datasets of ID images that could be used to reproduce and compare results~\cite{bulatov2022midv}.

\begin{figure}[ht]\captionsetup[subfigure]{font=scriptsize}
    \centering
    \begin{subfigure}[b]{0.37\textwidth}
        \centering
        \includegraphics[width=\textwidth]{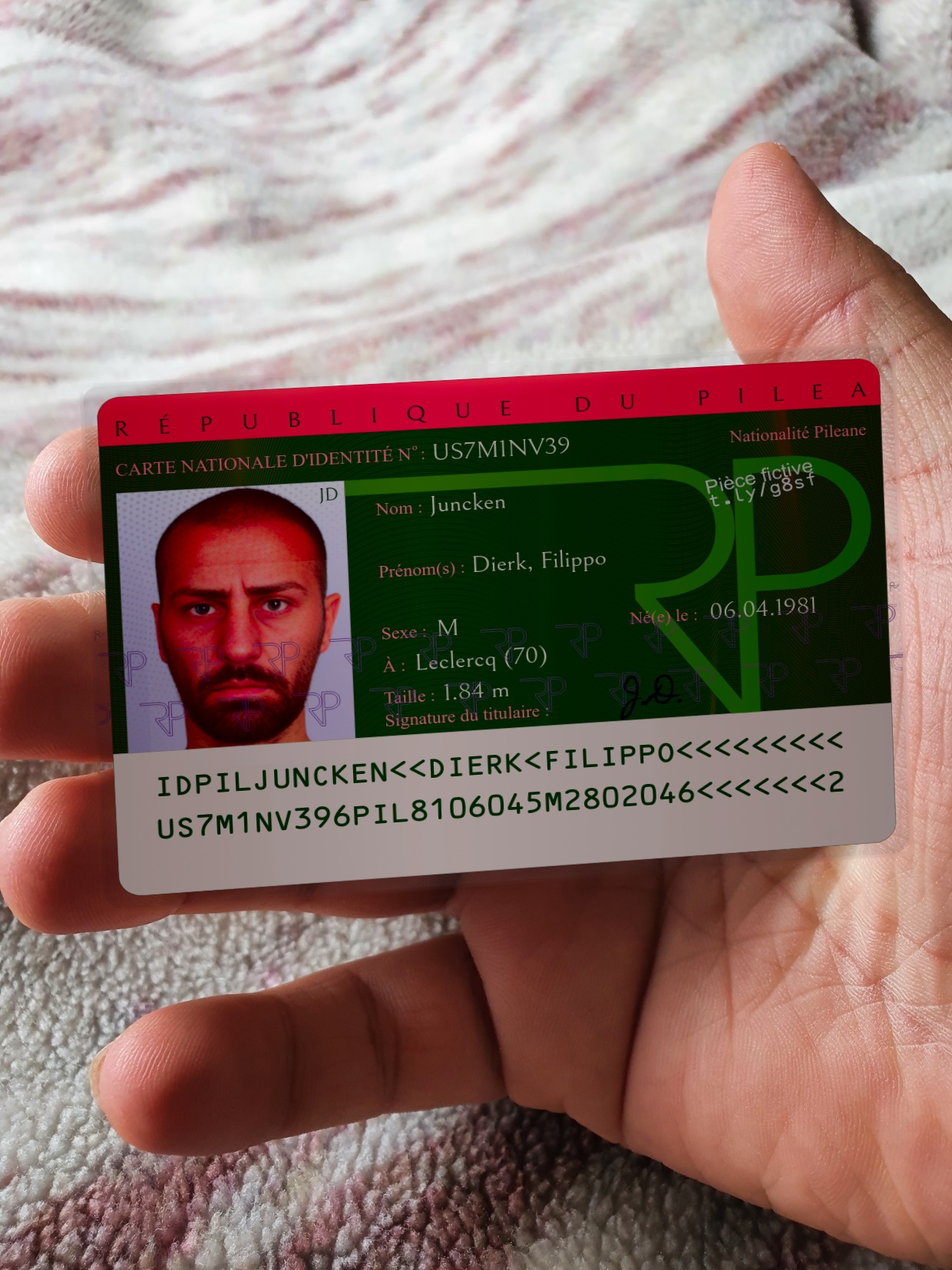}
        \caption*{ID\_CARD\_TD2\_A (front)}
    \end{subfigure}
    \hfill
    \begin{subfigure}[b]{0.37\textwidth}
        \centering
        \includegraphics[width=\textwidth]{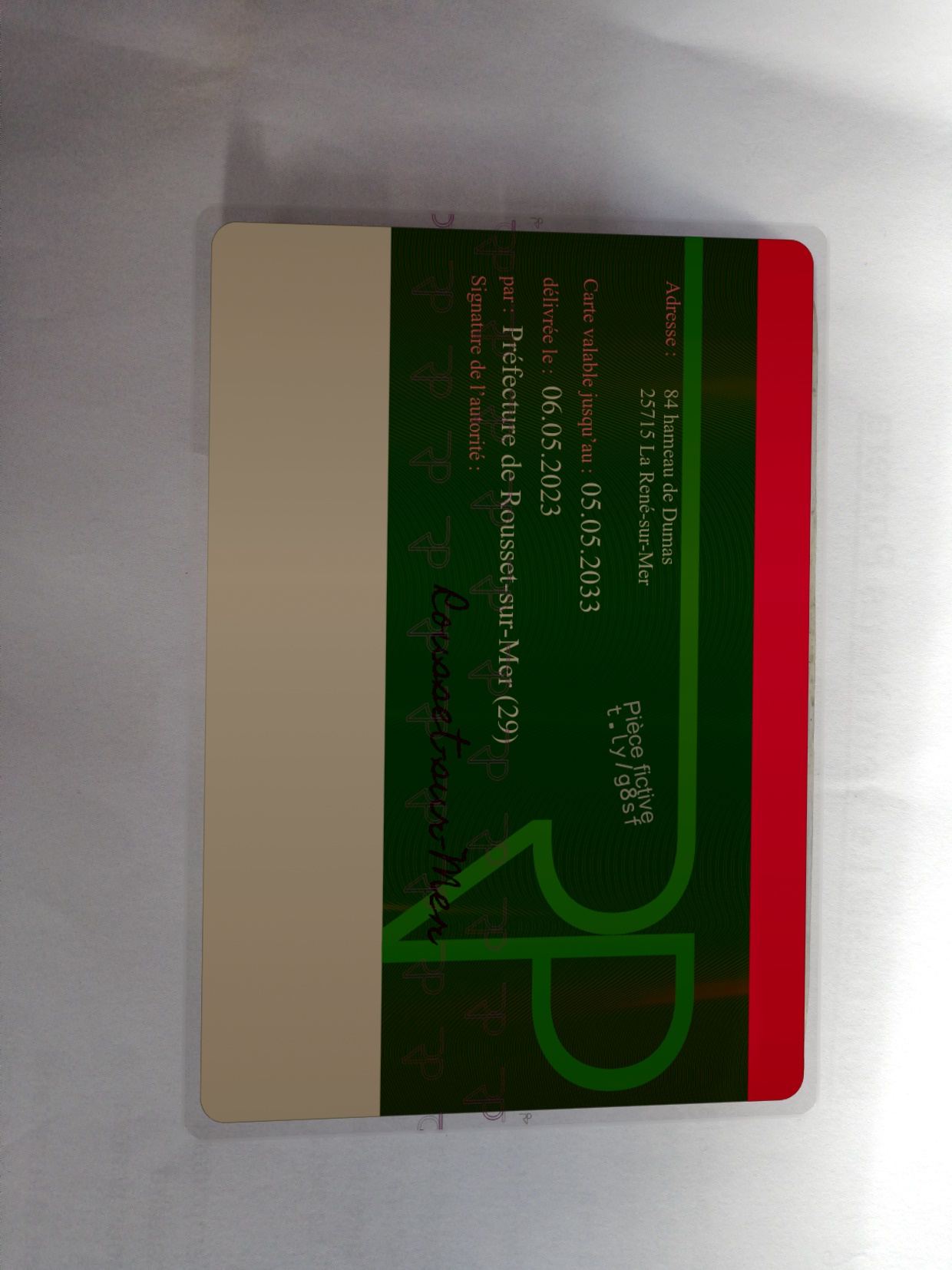}
        \caption*{ID\_CARD\_TD2\_A (back)}
    \end{subfigure}
    \hfill
    \begin{subfigure}[b]{0.22\textwidth}
        \centering
        \includegraphics[width=\textwidth]{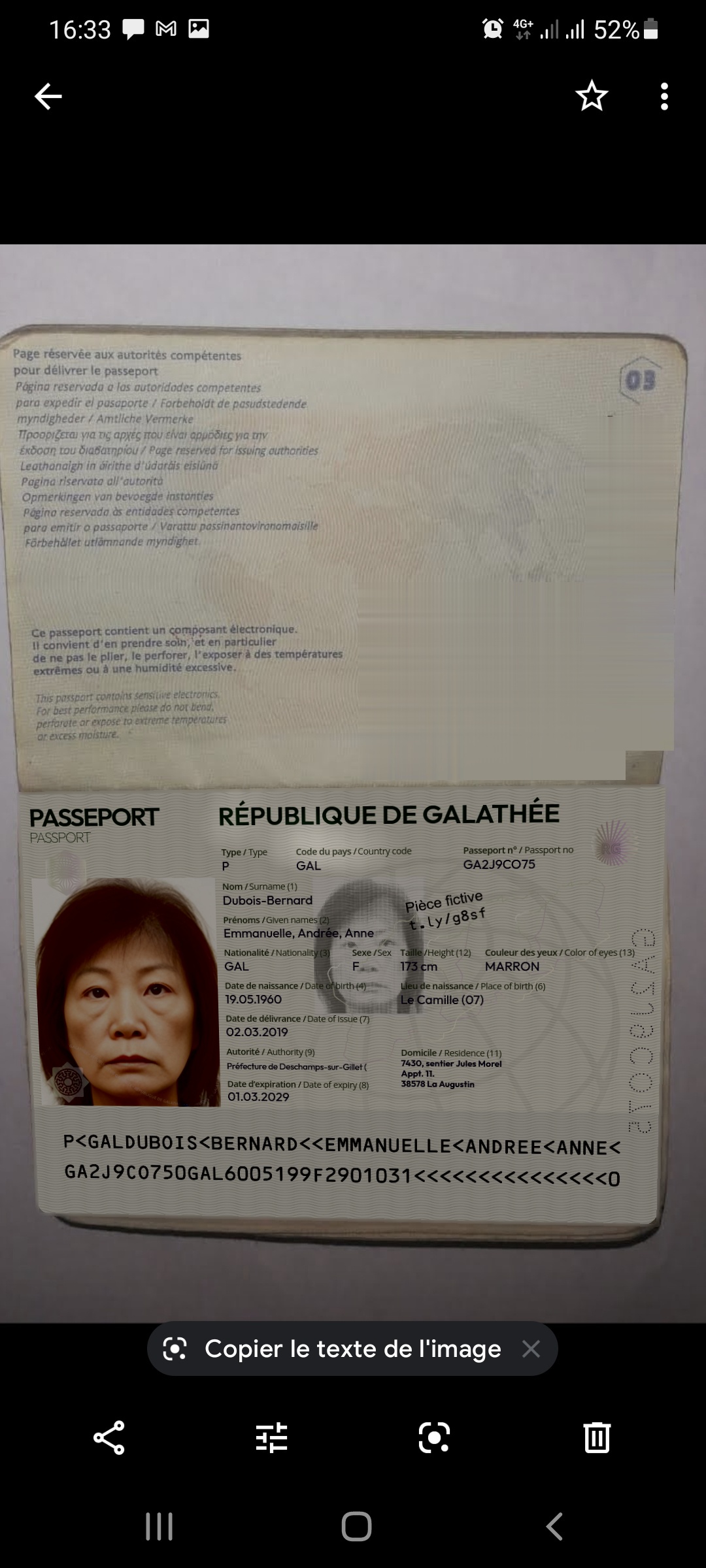}
        \caption*{PP\_TD3\_A}
    \end{subfigure}
    \\
    \begin{subfigure}[b]{0.33\textwidth}
        \centering
        \includegraphics[width=\textwidth]{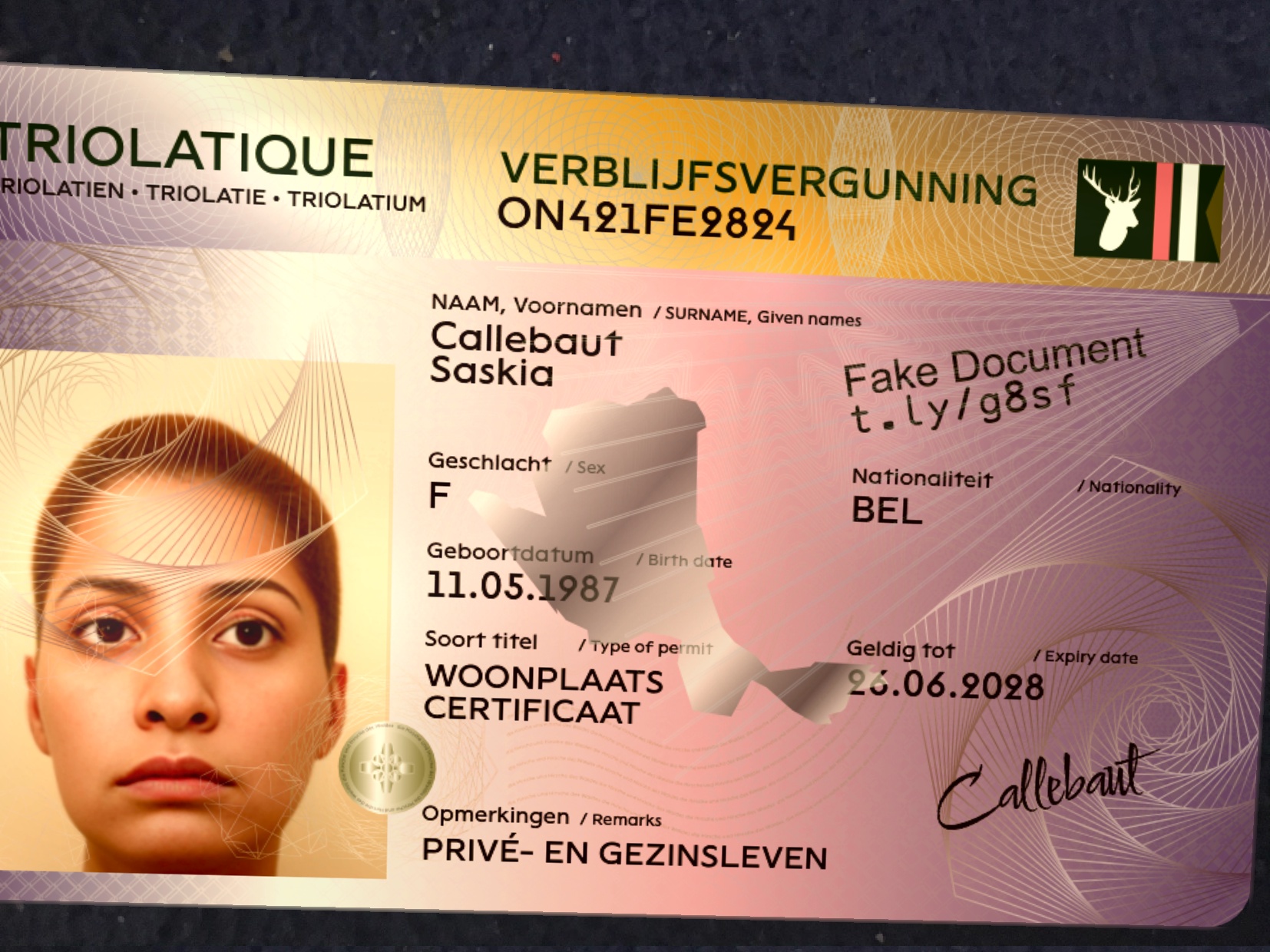}
        \caption*{RP\_CARD\_TD1 (front)}
    \end{subfigure}
    \hfill
    \begin{subfigure}[b]{0.31\textwidth}
        \centering
        \includegraphics[width=\textwidth]{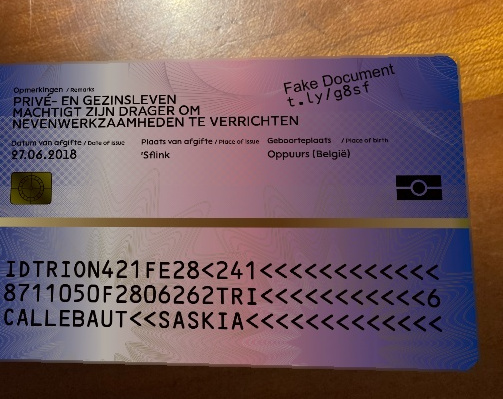}
        \caption*{RP\_CARD\_TD1 (back)}
    \end{subfigure}
    \hfill
    \begin{subfigure}[b]{0.32\textwidth}
        \centering
        \includegraphics[width=\textwidth]{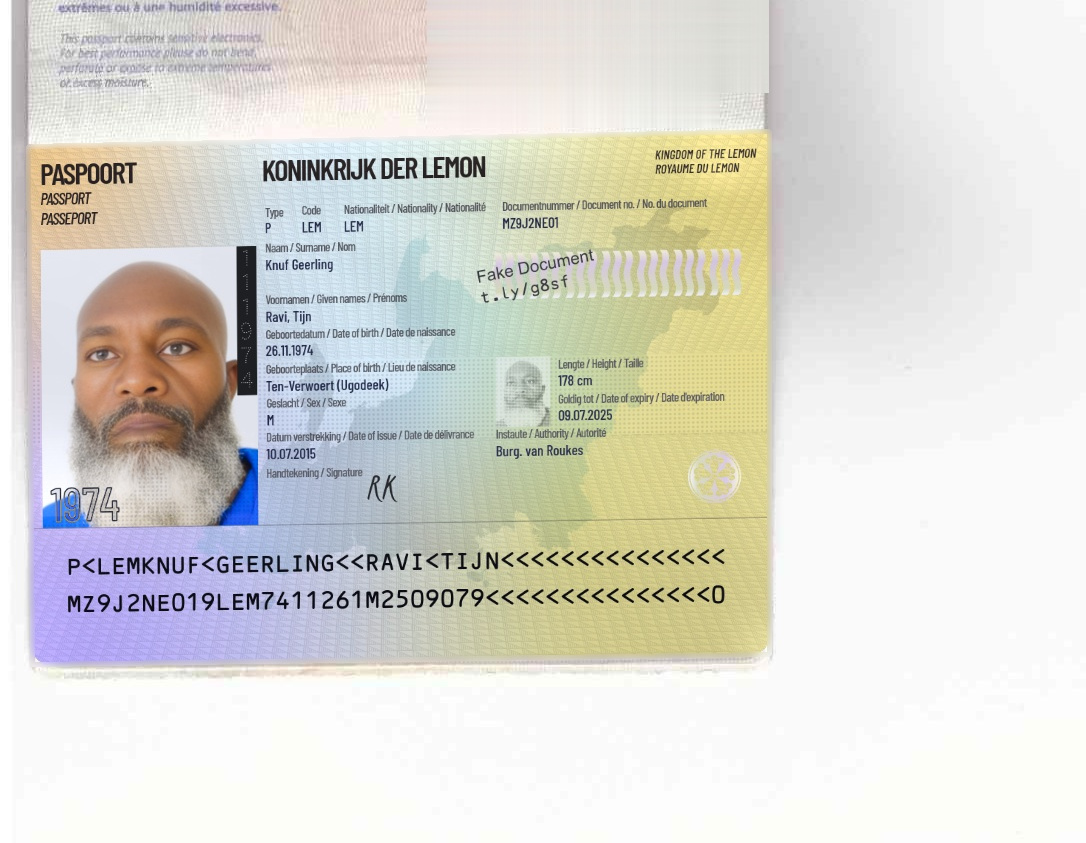}
        \caption*{PP\_TD3\_C (back)}
    \end{subfigure}
    \normalsize
    \caption{Six sample documents of the proposed DocXPand-25k dataset.}
    \label{fig:integrated_docs_samples}
\end{figure}

The main purpose of this paper is to present our ID images dataset, DocXPand-25k, consisting of 24\,994 richly labeled IDs images (a few samples are shown in Figure~\ref{fig:integrated_docs_samples}), which is suitable to evaluate methods on the following tasks :
\begin{itemize}
    \item ID classification,
    \item ID localization,
    \item detection of ID-specific features like identity photos (face detection), signatures and machine-readable zones (MRZ)
    \item ID text fields recognition.
\end{itemize}

\section{Related work}

In the past few years, the publication of the MIDV-family datasets~\cite{arlazarov2019midv, bulatov2020midv, bulatov2022midv, chernyshova2021midv} has had a significant impact in ID analysis fields. The MIDV-500 dataset~\cite{arlazarov2019midv}, containing 50 classes of IDs, was published in 2019. For each class, a specimen has been made using a laser printer and a laminator. The content of each of these specimens has been annotated, and each specimen was captured in video 10 times under varying visual conditions. This was the first public dataset to be focused on ID analysis and recognition, whether on image or video. Although, it has several limitations : every image has a unique resolution of 1920x1080 pixels and were taken using only two different smartphone cameras ; authors simulated uncontrolled environmental conditions by proposing five different environments for their video captures but the same five environments are used for each class; and the dataset is composed of only a single specimen per class of documents which vastly limits the interest of this dataset for OCR and classification evaluation.
In 2020, authors published supplementary materials to MIDV-500 dataset, named MIDV-2019~\cite{bulatov2020midv}, that addressed the lack of variability in capture conditions of the initial dataset, in terms of projective distortion, lighting conditions, etc.

Afterward, in 2022, they proposed MIDV-2020 dataset~\cite{bulatov2022midv}, composed of fake documents of 10 different classes (including Eastern, Northern and Southern European ID cards), containing 100 fake documents per class. The fake documents were printed on glossy photo paper, laminated, cut, and the corners were rounded to make them look like actual IDs. The text fields, signatures and photos have been artificially generated. For each document, a photo, two scans and a video were captured. A total of 72409 annotated images and video frames are publicly available. Compared to MIDV-500, this dataset partially solves the problem of the diversity of the proposed documents, since 100 different images per class are proposed. However, the authors don't provide tools nor protocol to generate more examples of each class. Also, some issues concerning the single photo resolution and the low diversity of acquisition devices used, as well as the limited number of capture environments, are still unsolved.

At last,~\cite{polevoy2022document} presented the DLC-2021 dataset, which consists of 1424 video clips captured in a wide range of real-world conditions, focused on tasks relating to ID forensics. It represents the same set of 10 ID classes as in MIDV-2020 dataset. The novelty of this dataset is that it contains frames from videos with color laminated fake IDs, color unlaminated copies, grayscale unlaminated copies, and screen captures of the documents.

Other datasets, with more limited impact on the literature, were also proposed. In 2018, LRDE researchers published a dataset~\cite{ngoc2018saliency} to evaluate their ID detection method based on saliency maps. It suffers from the same limitations as MIDV-500 (few samples per classes, low variability in the capture conditions). In 2020, the BID dataset~\cite{de2020bid} has been published, which focuses on Brazilian Identity Documents. This dataset is composed of about 29\,000 ID images, generated synthetically from 8 templates. This dataset suffers from a lack of realism, because the text printing technique is not very precise, and the templates were cleaned using inpainting. Also, the documents are straight and fill the images entirely, and the faces have been blurred, which prevents from using this dataset to evaluate document localization and face detection methods.

At the time of publication, we aren’t aware of any public dataset of ID images which combines all these features :
\begin{itemize}
    \item a large diversity of identities available per document class ;
    \item a large variability in acquisition conditions (diverse non-repeating environments, various acquisition devices and various lighting conditions) ;
    \item available tools to generate more images,
    \item and a rich enough labeling to enable benchmark classification, document localization, face detection, barcode detection, MRZ detection, and text field recognition methods.
\end{itemize}

In order to address these issues, we created a dataset of ID images, with the idea of producing high-quality, diverse and reusable data.

\section{Dataset}

\subsection{Study of real documents}

An ID is any document that can be used to prove the identity of its owner. It has to contain various details about the person, most commonly the person's full name, birth date, gender, photo and signature. Often, the document is also composed of an identification number and a MRZ. There is a wide diversity of IDs issued worldwide, and they can be organized using this information : the category of ID (identity card, passport, visa, residence permit, driving license, etc.), the country that issued the document and the period of issue of the document.

\begin{figure}[ht]\captionsetup[subfigure]{font=scriptsize}
    \centering
    \begin{subfigure}[b]{0.32\textwidth}
        \centering
        \includegraphics[width=\textwidth]{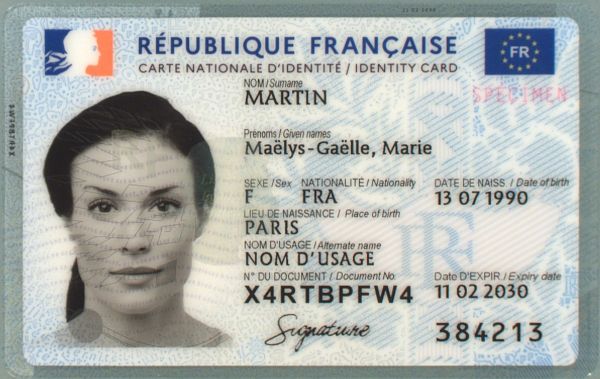}
        \caption*{FRA-BO-03001 (front)}
    \end{subfigure}
    \hfill
    \begin{subfigure}[b]{0.32\textwidth}
        \centering
        \includegraphics[width=\textwidth]{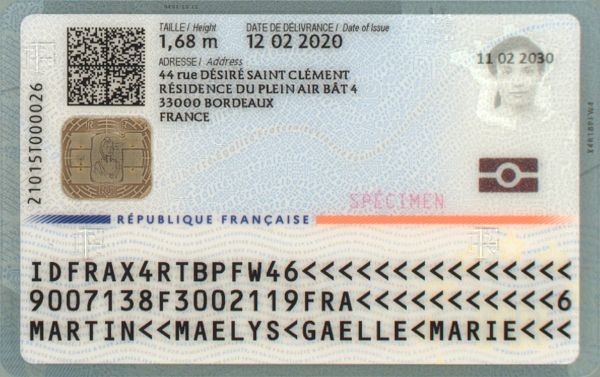}
        \caption*{FRA-BO-03001 (back)}
    \end{subfigure}
    \hfill
    \begin{subfigure}[b]{0.32\textwidth}
        \centering
        \includegraphics[width=\textwidth]{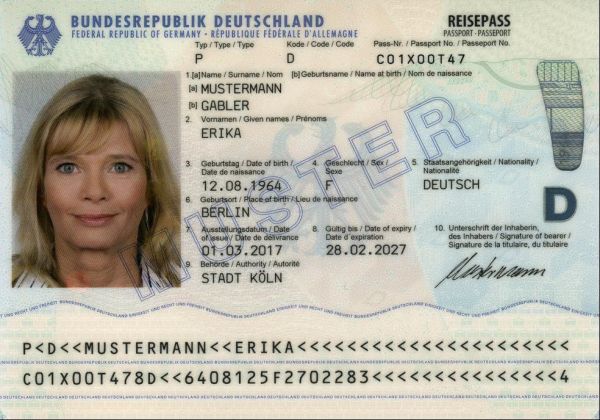}
        \caption*{DEU-AO-04001}
    \end{subfigure} \\
    \begin{subfigure}[b]{0.32\textwidth}
        \centering
        \includegraphics[width=\textwidth]{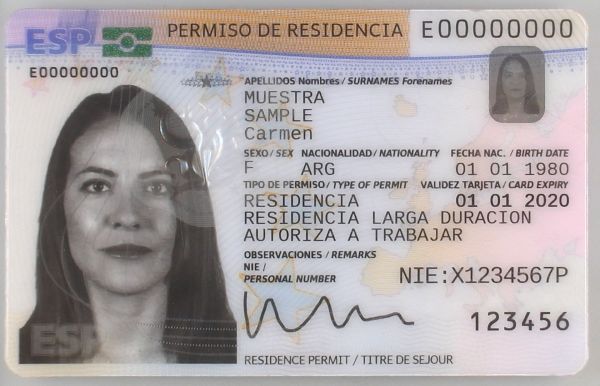}
        \caption*{ESP-HO-03001 (front)}
    \end{subfigure}
    \hfill
    \begin{subfigure}[b]{0.32\textwidth}
        \centering
        \includegraphics[width=\textwidth]{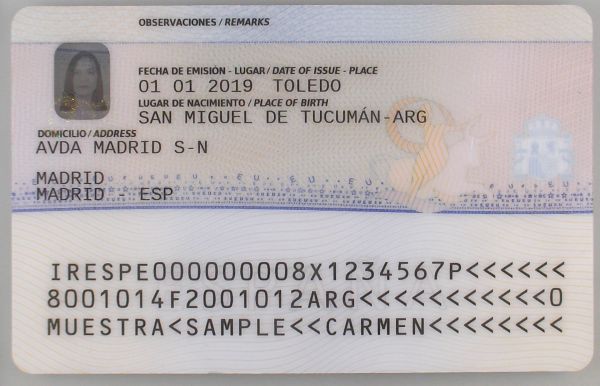}
        \caption*{ESP-HO-03001 (back)}
    \end{subfigure}
    \hfill
    \begin{subfigure}[b]{0.32\textwidth}
        \centering
        \includegraphics[width=\textwidth]{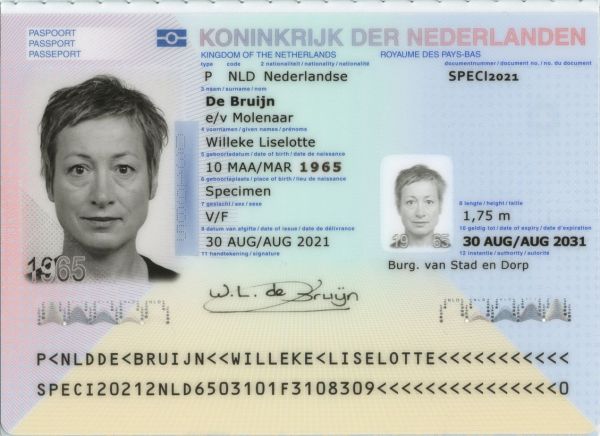}
        \caption*{NLD-AO-05001}
    \end{subfigure}
    \caption{Specimens of a few ID classes. Top line, left to right: French identity card (front and back sides), German passport (bio-data page). Bottom line, left to right: Spanish residence permit (front and back sides), Dutch passport (bio-data page).}
    \label{fig:id_docs_specimens}
\end{figure}

We call a class of IDs a set of documents that share the same category, country and period of issue. Generally, IDs of the same class are very close visually, since they are issued based on the same fixed layout. Some examples of IDs are presented in Figure~\ref{fig:id_docs_specimens}. The online knowledge base known as PRADO~\cite{prado2022}, published by the Council of the European Union, is a good reference to study the IDs classes, their visual layout and their security features.

Many visual, technical and security aspects of a wide variety of travel documents have been standardized by the International Civil Aviation Organization since the 1980s, through the publication of Doc 9303~\cite{doc9303}. Among these standards, the use of the following Machine-Readable Official Travel Documents formats has thus become ubiquitous : Type 1 (TD1), Type 2 (TD2) and Type 3 (TD3). TD1 and TD2 documents are often used for identity cards, residence permits and driving licenses, while TD3 is typical of passports. Their physical sizes follow the specifications of ISO/IEC 7810 ID1, ID2 and ID3 formats.

\subsection{Vectorial templates}

\begin{figure}[ht]\captionsetup[subfigure]{font=scriptsize}
    \centering
    \begin{subfigure}[t]{0.32\textwidth} 
        \centering
        \includegraphics[width=\textwidth]{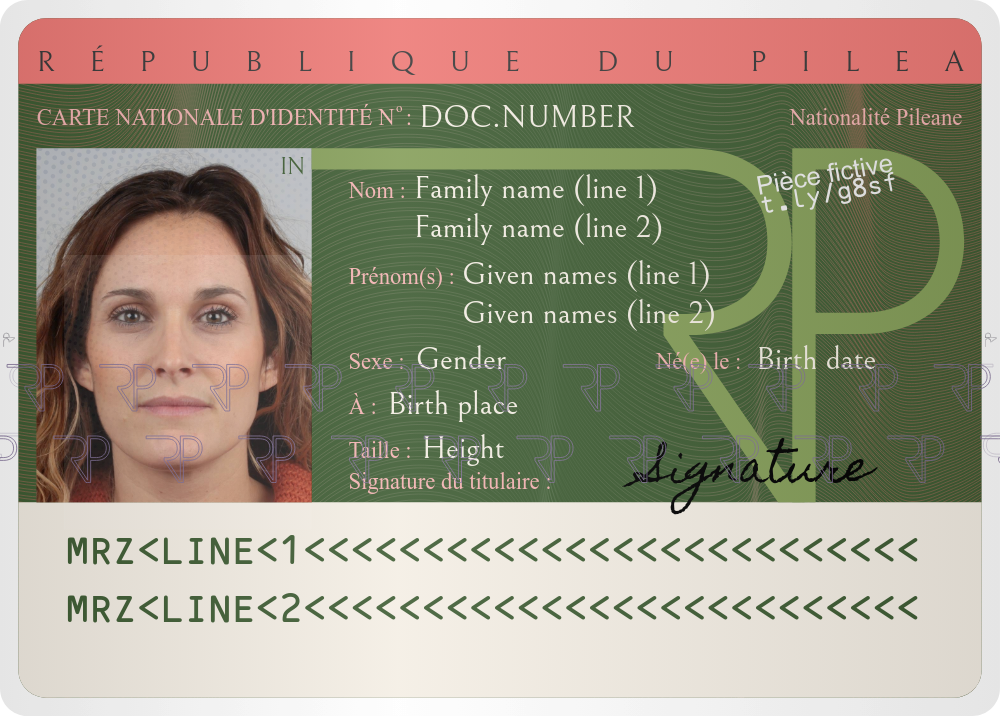}
        \caption*{ID\_CARD\_TD2\_A (front)} 
    \end{subfigure}
    \hfill 
    \begin{subfigure}[t]{0.32\textwidth}
        \centering
        \includegraphics[width=\textwidth]{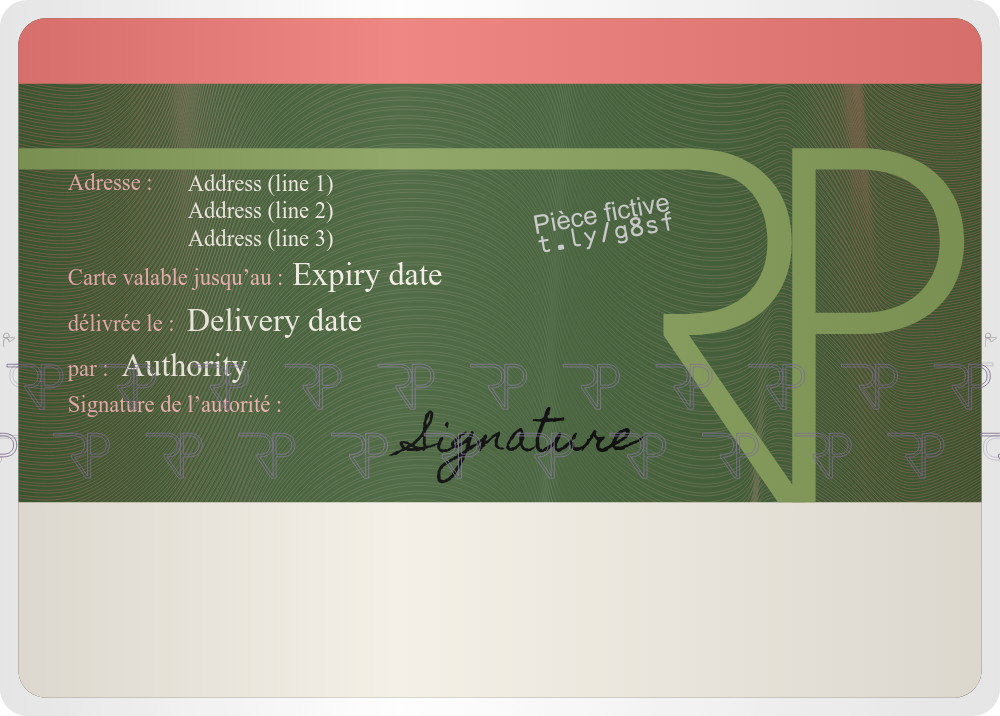}
        \caption*{ID\_CARD\_TD2\_A (back)}
    \end{subfigure}
    \hfill
    \begin{subfigure}[t]{0.32\textwidth}
        \centering
        \includegraphics[width=\textwidth]{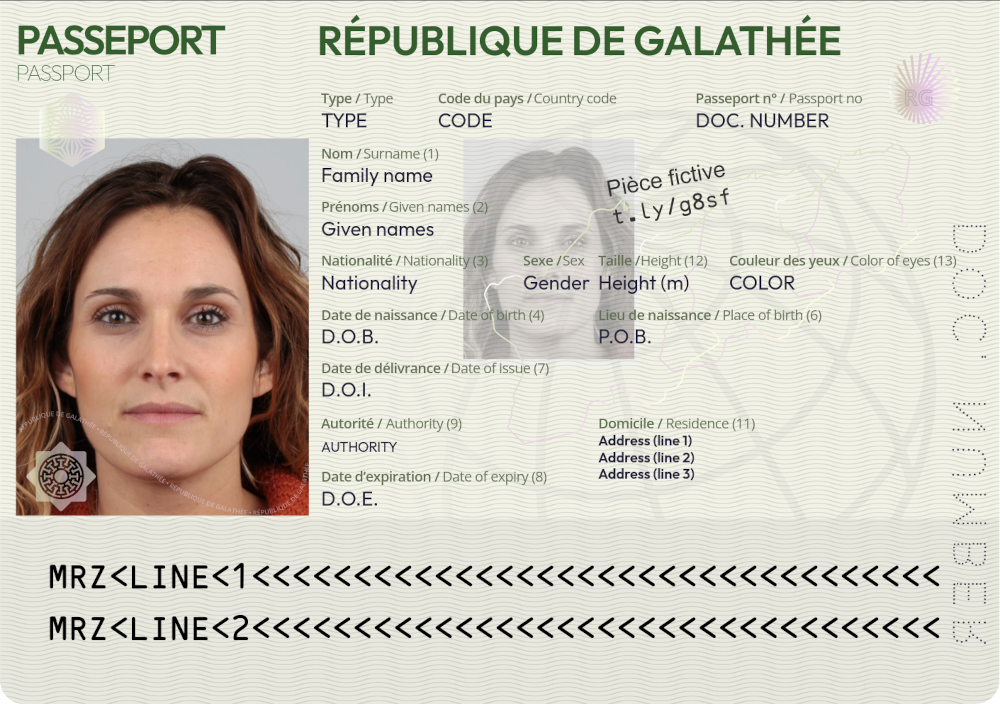}
        \caption*{PP\_TD3\_A}
    \end{subfigure}
    \\
    \begin{subfigure}[t]{0.32\textwidth}
        \centering
        \includegraphics[width=\textwidth]{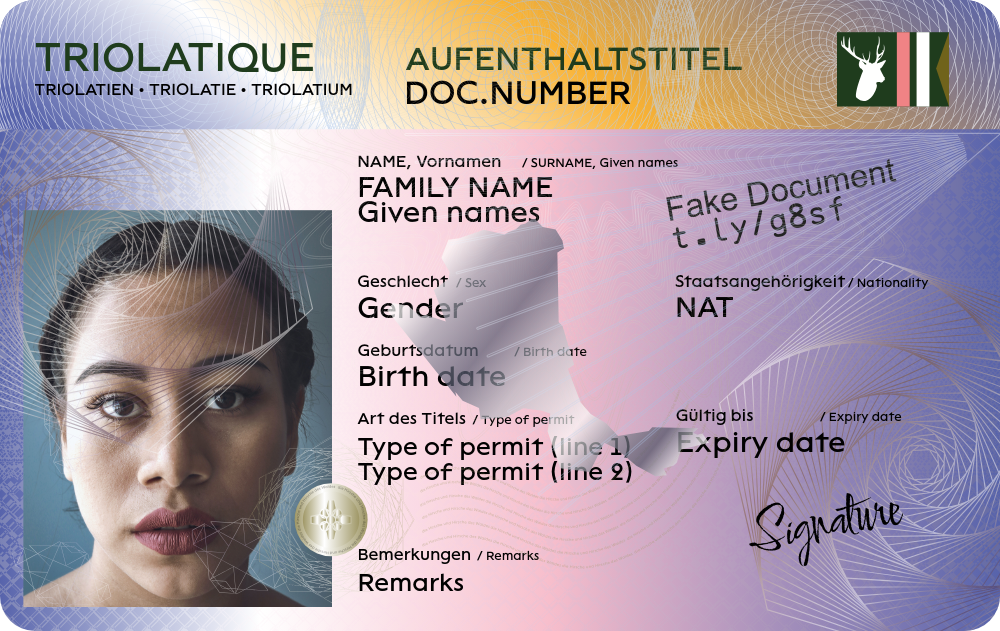}
        \caption*{RP\_CARD\_TD1 (front)}
    \end{subfigure}
    \hfill
    \begin{subfigure}[t]{0.32\textwidth}
        \centering
        \includegraphics[width=\textwidth]{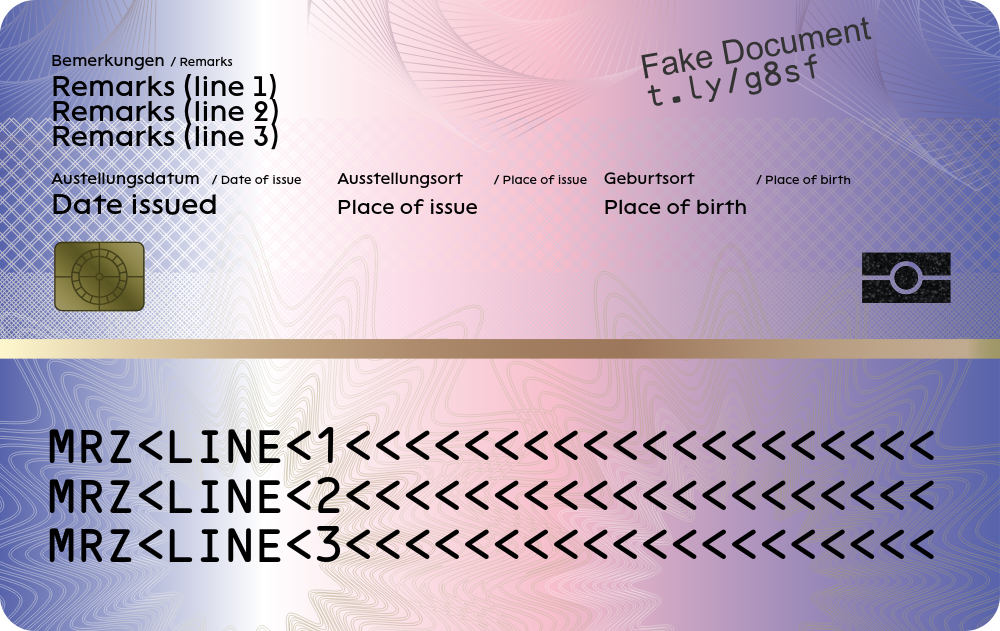}
        \caption*{RP\_CARD\_TD1 (back)}
    \end{subfigure}
    \hfill
    \begin{subfigure}[t]{0.32\textwidth}
        \centering
        \includegraphics[width=\textwidth]{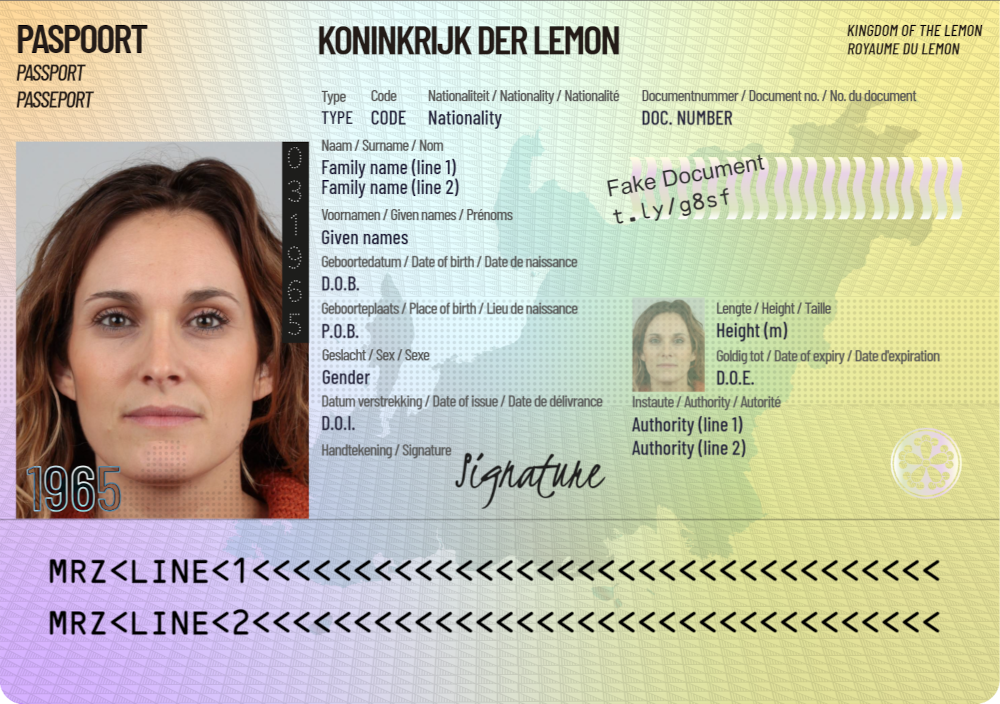}
        \caption*{PP\_TD3\_B}
    \end{subfigure}
    \caption{Some templates designed for DocXPand.}
    \label{fig:templates}
\end{figure}

The DocXPand-25k dataset comprises nine fictitious document classes, including four classes of identity cards, 2 classes of residence permits, and three classes of passports. For ability reasons, we chose to design templates of documents that could be issued by fictitious states speaking west-european languages only. We hired a professional graphic designer to design vectorial templates for four identity cards (both front and back sides), two residence permits (again, front and back sides), and three passports based on detailed specifications we wrote. Some of these templates are shown in Figure~\ref{fig:templates}.

\subsection{Document image generation}

\begin{figure}[ht]
    \centering
    \includegraphics[width=1.0\textwidth]{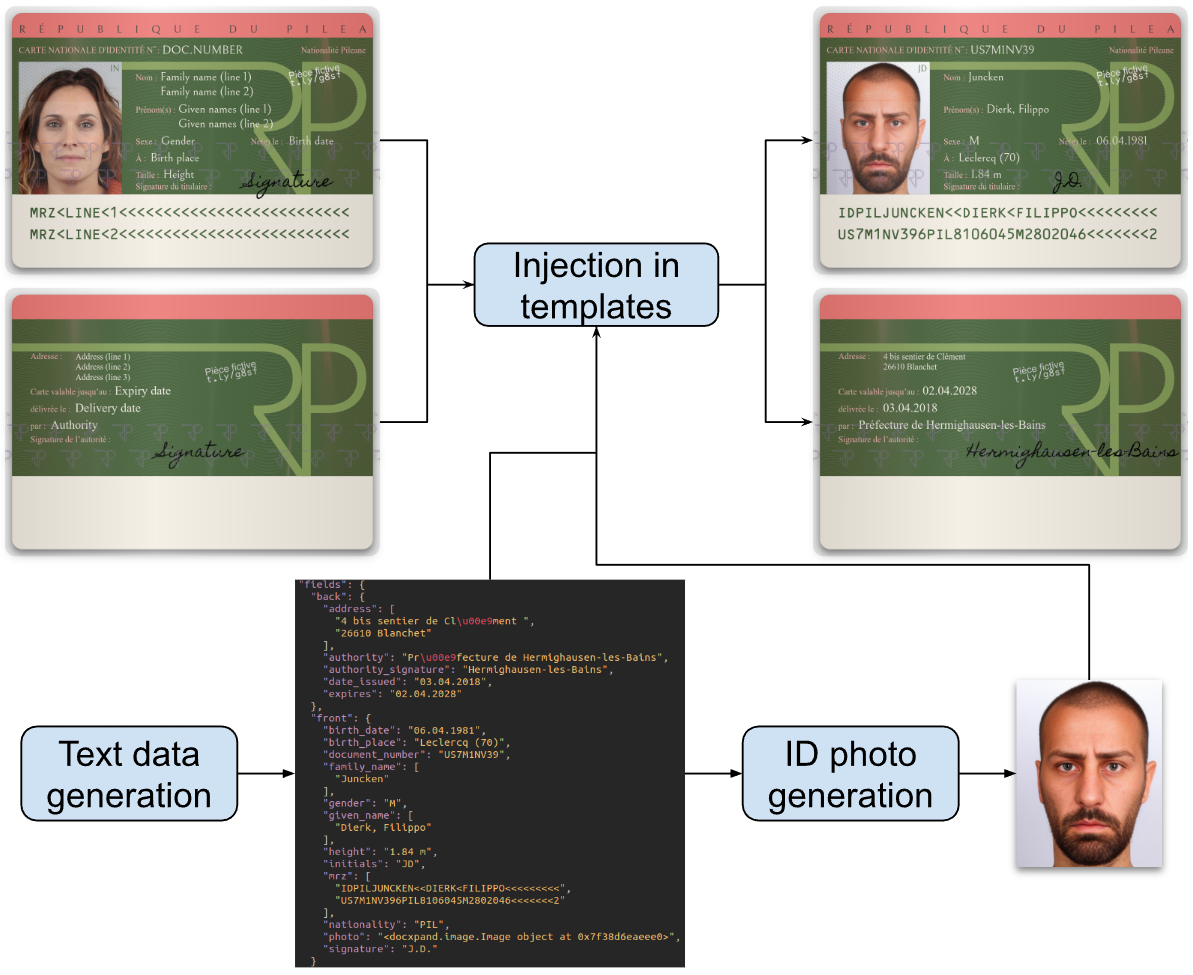}
    \caption{Document image generation process.}
    \label{fig:document_generation_process}
\end{figure}

Our vectorial templates have been manually labeled to indicate exactly where the actual text and image contents must be placed. We created dedicated text and image generators that produce the content for each field. In order to avoid any personal data issues, the personal fields generator does not use any existing identity database, but samples fake values randomly from some distributions. The address generator also creates entirely fake values for every address level (in particular street, postcodes and city). We have built these generators over the \texttt{faker}\footnote{https://github.com/joke2k/faker} python tool. For some templates, we also generate data matrix and barcodes using \texttt{zxingcpp}\footnote{https://github.com/zxing-cpp/zxing-cpp}. Finally, we have used Stable Diffusion v1.5~\cite{rombach2022high} to generate identity photos of faces, by using txt2img API from Stable Diffusion WebUI \footnote{https://github.com/AUTOMATIC1111/stable-diffusion-webui}, with a few source photos to control the pose and a prompt (see Listing~\ref{lst:prompt}) depending on the generated text fields (age and gender). To mitigate racial bias in face generation, we also sampled the ethnic origins from a uniform distribution of 10 manually selected origins. For each document to generate, the generated text fields, photos and barcodes are injected at the target position on the template, as shown in Figure~\ref{fig:document_generation_process}.

\begin{code}
    \captionof{listing}{Prompt template for identity photos generation}
    \label{lst:prompt}
    \begin{minted}[numbersep=5pt,
               frame=lines,
               framesep=2mm]{python}
prompt_template = (
    "face of a {gender}, {ethnicity}, {age} years old, neutral "
    "expression, body facing the camera, centered face, "
    "eyes looking front to the camera, clear background, "
    "color photography"
)
negative_prompt = (
    "(((text))), (((watermark))), (((black and white))), "
    "glasses, hat, jewelry, smiling, naked"
)
\end{minted}
\end{code}

\subsection{Document image integration in backgrounds}

\begin{figure}[ht]
    \centering
    \includegraphics[width=1.0\textwidth]{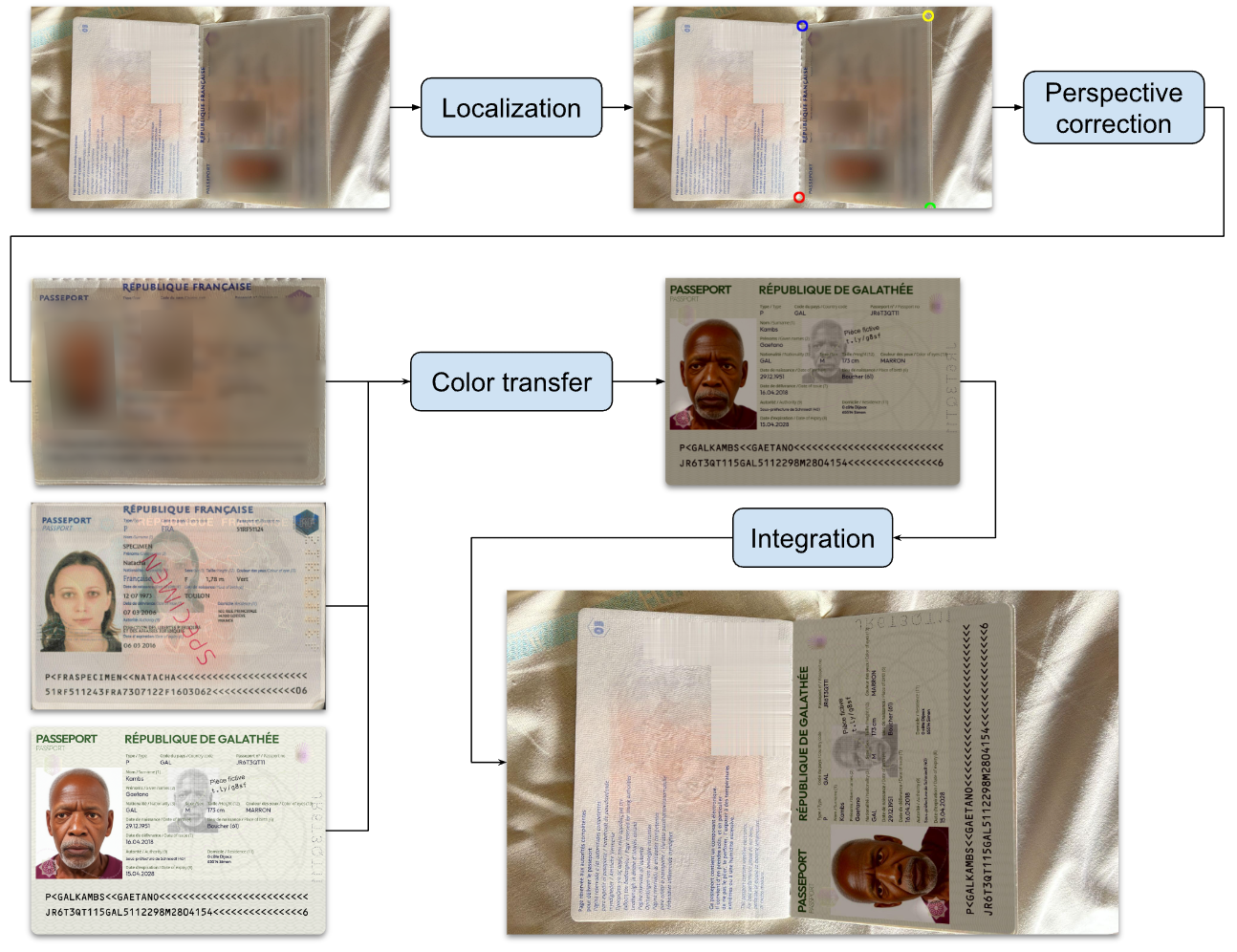}
    \caption{Document image integration in backgrounds process.}
    \label{fig:document_integration_process}
\end{figure}

We collected internally about 3.2k images (photos, scans and screenshots) containing real TD1 and TD2 documents, and about 2.6k images containing real French passports. These images serve as backgrounds in our generation process, where the original documents are replaced by our generated ones. To this end, we used SDL-Net~\cite{kabeshova2024sdlnet}, a corner-based document localization model that we trained internally on other images coming from the production data of our platform, to localize the 4 labeled corners (top-left, top-right, bottom-left and bottom-right) of the documents inside these images. The real IDs are compared to their specimen images, like those presented in Figure~\ref{fig:id_docs_specimens}, to estimate a color transformation by estimating luminosity and color shift in L*a*b* color space. This transformation is then applied to the generated document image using the transfer method described in~\cite{reinhard2001color}, to make the integration more realistic and to add luminosity and color variations in the dataset. We also developed an algorithm to transfer flashes, light spots, and shadows to the generated document image. Finally, the transformed generated document image is integrated to the background, replacing the original document. This integration process is illustrated in Figure~\ref{fig:document_integration_process}, and some result images (corresponding to templates shown in Figure~\ref{fig:templates}) are shown in Figure~\ref{fig:integrated_docs_samples}.

For all real documents in our background images, we added a small margin to maximize the probability of covering the entire document. For French passports images specifically, we also inpainted the signature and the document number from the upper part (page 3) with a custom inpainting algorithm. We took extra care to include no personal information from the original documents in our dataset, by manually reviewing each final image twice (by two different annotators), to remove images that could leak any personal information.

This whole image generation process differentiates our contribution from MIDV-2020~\cite{bulatov2022midv}, because it allows generating an arbitrary number of images for each document class without much manual effort, and also increases the variability of backgrounds, of acquisition devices and acquisition conditions.

\subsection{Dataset format and labels}

In addition to the generated document images, we provide JSON files that describe the dataset and provide rich labels for each image, which include :
\begin{itemize}
    \item classification label ;
    \item coordinates of the ID in the image ;
    \item coordinates of the identity photo, ghost image and barcode, when relevant ;
    \item and coordinates and exact value of each text field.
\end{itemize}

\subsection{Statistics}

The proposed dataset includes a total of 24\,994 labeled document images, distributed among 15 classes (9 templates including 6 two-sided templates and 3 single-sided ones), and 237\,895 labeled text fields, including single-line and multi-line fields. More statistical details are presented in Table~\ref{tab:statistics}.

\begin{table}[ht]
    \begin{adjustbox}{width=\textwidth}
        \begin{tabular}{|c|c|c|c|c|c|}
            \hline
            \textbf{Template}                            & \textbf{Side} & \textbf{\# images} & \textbf{\# text fields} & \textbf{\# photos} & \textbf{\# barcodes} \\ \hline
            \multirow{2}{*}{ \texttt{ID\_CARD\_TD1\_A} } & front         & 1\,837              & 18\,370                  & 1\,837              & \textit{NA}          \\ \cline{2-6}
                                                         & back          & 1\,836              & 9\,180                   & 1\,836              & 1\,836                \\ \hline
            \multirow{2}{*}{ \texttt{ID\_CARD\_TD1\_B} } & front         & 1\,857              & 24\,141                  & 3\,714              & \textit{NA}          \\ \cline{2-6}
                                                         & back          & 1\,846              & 12\,922                  & 1\,846              & \textit{NA}          \\ \hline
            \multirow{2}{*}{ \texttt{RP\_CARD\_TD1} }    & front         & 1\,858              & 20\,438                  & 1\,858              & \textit{NA}          \\ \cline{2-6}
                                                         & back          & 1\,861              & 9\,305                   & \textit{NA}        & \textit{NA}          \\ \hline
            \multirow{2}{*}{ \texttt{ID\_CARD\_TD2\_A} } & front         & 1\,784              & 19\,624                  & 1\,784              & \textit{NA}          \\ \cline{2-6}
                                                         & back          & 1\,791              & 8\,955                   & \textit{NA}        & \textit{NA}          \\ \hline
            \multirow{2}{*}{ \texttt{ID\_CARD\_TD2\_B} } & front         & 1\,800              & 18\,000                  & 3\,600              & \textit{NA}          \\ \cline{2-6}
                                                         & back          & 1\,733              & 5\,199                   & \textit{NA}        & 1\,733                \\ \hline
            \multirow{2}{*}{ \texttt{RP\_CARD\_TD2} }    & front         & 1\,241              & 14\,892                  & 1\,241              & \textit{NA}          \\ \cline{2-6}
                                                         & back          & 1\,236              & 4\,944                   & 1\,236              & \textit{NA}          \\ \hline
            \multicolumn{2}{|c|}{ \texttt{PP\_TD3\_A } } & 1\,473         & 25\,041             & 2\,946                   & \textit{NA}                               \\ \hline
            \multicolumn{2}{|c|}{ \texttt{PP\_TD3\_B } } & 1\,413         & 22\,608             & 1\,413                   & \textit{NA}                               \\ \hline
            \multicolumn{2}{|c|}{ \texttt{PP\_TD3\_C } } & 1\,428         & 24\,276             & 2\,856                   & \textit{NA}                               \\ \hline
            \multicolumn{2}{|c|}{ \textbf{TOTAL} }       & 24\,994        & 237\,895            & 26\,167                  & 3\,569                                     \\ \hline
        \end{tabular}
    \end{adjustbox}
    \normalsize
    \caption{Statistics of the proposed dataset.}
    \label{tab:statistics}
\end{table}

\subsection{Visual similarity}

We aim at creating a benchmark dataset that represents well the diversity of real images processed by industrial ID analysis systems. To quantify this, we propose using LPIPS~\cite{zhang2018unreasonable} to measure the visual similarity between a private reference dataset, named $R$ hereafter, of 2\,719 images of real IDs captured by end-users of our industrial platform, and five benchmark datasets: MIDV-500, MIDV-2019, MIDV-2020 and DLC-2021 and DocXPand-25k (ours), named $D_i, i \in [1, 5]$ hereafter. None of the images of $R$ were used to generate images of $D_5$ (ours).

\begin{figure}[ht]
    \centering
    \includegraphics[width=\textwidth]{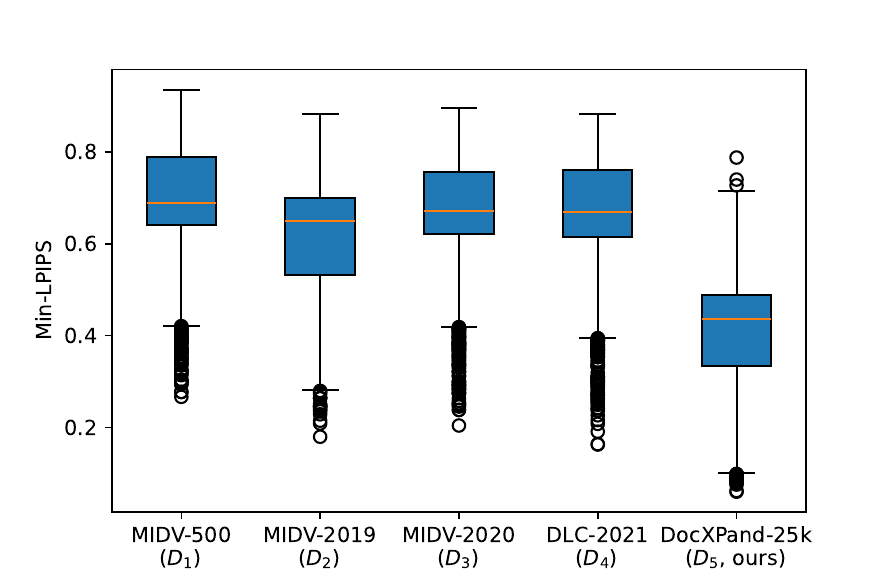}
    \caption{Box-plot of min-LPIPS distributions between a private reference dataset and five benchmark datasets.}
    \label{fig:lpips}
\end{figure}

For each of $D_i, i \in [1, 4]$, we randomly selected 3\,000 frames, and for $D_5$, we simply took the 3\,000 images from the test dataset. We used a pre-trained model\footnote{\href{https://github.com/richzhang/PerceptualSimilarity}{https://github.com/richzhang/PerceptualSimilarity}} to compute similarity matrices between $R$ and $D_i, i \in [1, 5]$. For each sample of $R$, we find the sample of $D_i, i \in [1, 5]$ minimizing the LPIPS value. The distributions of minimal-LPIPS per benchmark dataset $D_i$ is presented in Figure~\ref{fig:lpips}. The obtained results shows that our proposed dataset is perceptually closer to document images typically processed by industrial platforms like ours.

\section{Metrics and baseline experiments}
\label{sec:baselines}

In this section, we propose evaluation metrics and simple baselines for the ID classification, ID localization and text field recognition tasks.

\subsection{ID classification}
\label{sec:id_classification}
Document classification is a sub-field of image classification, consisting generally in assessing the class of the document present in an image using visual and/or semantic aspects. In template-based document analysis systems, it is a crucial step because it allows selecting the right template for each image.

For the purpose of this baseline, we have split our dataset into three subsets: training, validation and test. We have paid attention to not share identical backgrounds between these subsets, so that all images generated using the same background image are all in the same subset. The training subset contains 18\,153 labeled document images, while the validation and test subsets contains 3\,395 and 3\,446 documents labeled document images respectively (between 200 and 263 images per template, and 228 per template on average).

To assess the quality of a document classification algorithm $C$, we use the classification accuracy metric, defined as the proportion of documents correctly classified by $C$ over the total number of documents in a set $D$ :
\begin{equation}
    \textrm{acc}_{\textrm{cls}}(C, D) = \frac{\textrm{card}(\{d \in D, C(d) = \tilde d_c)}{\textrm{card}(D)},
\end{equation}
where $\tilde d_c$ is the ground-truth class of document $d$.

Much research has been conducted in the field of image classification since the publication of the ImageNet benchmark~\cite{deng2009imagenet}, and most of the proposed methods use deep learning models. We provide baselines using a few popular and lightweight CNN architectures targeting mobile devices, yet not state-of-the-art : MobileNet-V3~\cite{howard2019searching} (\texttt{Small} and \texttt{Large} variations) and EfficientNet-V2\cite{tan2021efficientnetv2} (\texttt{B0} lightweight variation). Test metrics are presented in Table~\ref{tab:classification_baselines}.

\begin{table}[ht]
    \centering
    \begin{tabular}{|c|c|c|c|}
        \hline
        \textbf{Model}               & \textbf{\# params} & $\textbf{acc}_{cls}$ & throughput    \\ \hline
        MobiletNet-V3 \texttt{Small} & 2.9M               & 99.65\%              & \textbf{97.9} \\ 
        MobiletNet-V3 \texttt{Large} & 5.4M               & \textbf{99.80\%}     & 80.2          \\ 
        EfficientNet-V2 \texttt{B0}  & 7.4M               & 99.16\%              & 71.7          \\ \hline
    \end{tabular}
    \caption{Classification metrics on test set (3\,446 images, 15 classes)}
    \label{tab:classification_baselines}
\end{table}

For each model, images are resized to 224x224 pixels in training and inference phase, but no prior document localization is performed. Models have been trained on an NVidia Titan X, with Adam optimizer ($lr=3e-4$) and data augmentation (random contrast, saturation, rotation, crop). Throughput is expressed in images/sec, and estimated using ONNX Runtime inference on Intel i9-12900H CPU. We clearly see that lightweight CNN models are sufficient for industrial purpose on this classification task, and no major improvement are expected from larger CNN or transformer-based model.

\subsection{ID localization}
\label{sec:id_localization}
Document localization is the process of finding the position and shape of a document in an image, which is an important pre-processing step in structured document analysis systems before extracting information from text and image fields.
We consider that the documents of our dataset can be represented as quadrilaterals, even if some have rounded angles, that can be defined using the coordinates of the four corners : top left, top right, bottom right and bottom left.

To assess the quality of a document localization algorithm $L$ on a document $d$, we use the Intersection over Union (IoU) metric : 
\begin{equation}
    \textrm{IoU}(L, d) = \frac{ \textrm{area} ( L(d) \cap \tilde d_l ) }{ \textrm{area} ( L(d) \cup \tilde d_l ) },
\end{equation}
where $\tilde d_l$ is the ground-truth quadrilateral representing the position of the document $d$ in the image.

We consider, like~\cite{bulatov2022midv}, that a document is well localized by $L$ if $\textrm{IoU}(L,d) \geq 0.9$.
The localization accuracy can be defined as the proportion of documents well localized over the total number of documents in a set $D$ :
\begin{equation}
    \textrm{acc}_{\textrm{loc}}(L, D) = \frac{\textrm{card}(\{d \in D, \textrm{IoU}(L, d) \geq 0.9\})}{\textrm{card}(D)}
\end{equation}

The main categories of approaches used to detect documents on images are based on corner detection~\cite{zhu2019coarse}, edge detection~\cite{tropin2021advanced} or segmentation~\cite{ngoc2019document}. Here, we provide a baseline using SDL-Net~\cite{kabeshova2024sdlnet}, a corner-based document localization model. We have used the same dataset split as the one described in section \ref{sec:id_classification}.  We report an average IoU of $0.969\pm0.096$, with an $\textrm{acc}_{\textrm{loc}} = 95.4\%$. We have trained the model for 180 epochs, during approximately 6 days, on an NVidia Titan X, . The results of this baseline seem rather decent regarding the difficulty of the task and results of concurrent methods on similar benchmarks~\cite{zhu2019coarse, tropin2021advanced}.

\subsection{Text fields recognition (OCR)}
\label{sec:text_fields_recognition}

In this section, we focus on the text fields recognition problem, where we assume that the IDs have already been located, and divided into zones in the image, each zone representing a different text field to recognize using an OCR.

To assess the quality of a text recognition algorithm $T$ on a text field $f$, we suggest using the widely used Character Error Rate (CER) metric, which is based on the Levenshtein edit-distance :
\begin{equation}
    \textrm{CER}(T, f) = \frac{ \textrm{levenshtein} (T(f), \tilde f) }{ \textrm{len} ( \tilde f ) },
\end{equation}
where $\tilde f$ is the ground-truth text contained in the field $f$. For multi-line fields, the CER is computed by treating the new line as a regular character, both in ground-truth and predicted text.
We consider that a text field is well recognized by $T$ if $\textrm{CER}(T,f) = 0$, which is equivalent to $T(f) = \tilde f$. The recognition accuracy can then be defined as the proportion of text fields well recognized over the total number of text fields in a set $F$ :
\begin{equation}
    \textrm{acc}_{\textrm{rec}}(T, F) = \frac{\textrm{card}(\{f \in F, \textrm{CER}(T, f) = 0\})}{\textrm{card}(F)}
\end{equation}

The text fields of our dataset belong to four distinct categories: numerical fields (such as numeric dates, document numbers and other identifiers), alpha fields (such as first names and last names), alphanumeric fields (such as addresses, mixed birth date and place fields) and machine readable zones (MRZ).
For the purpose of this baseline, we have split our dataset into three subsets: training, validation and test. We have followed a specific constraint for two-sided templates (ID cards and residence permits): in the validation and test subsets, both sides of each document are always present (which could be useful to evaluate front and back field extraction consistency). However, in the training subset, some documents only have one side because the other was rejected during our manual review process.
The training subset contains 18\,994 labeled document images (classes are not perfectly balanced), while the validation and test subsets include exactly 200 labeled document images per class for a total of 3\,000 images each.

We provide a baseline of text field recognition using Tesseract OCR~\cite{smith2007overview} version 5.3.3, shipped with \texttt{tesserocr}\footnote{\href{https://github.com/sirfz/tesserocr}{https://github.com/sirfz/tesserocr}} version 2.6.2, and using page segmentation mode 6, \texttt{"SINGLE\_BLOCK"}, to treat inputs as blocks of text with one or more lines. The list of installed languages to process the input field images are: Catalan, Dutch, English, French, German, Italian, Portuguese and Spanish. A model specifically trained on OCR-B font\footnote{\texttt{best} model from \href{https://github.com/Shreeshrii/tessdata_ocrb}{https://github.com/Shreeshrii/tessdata\_ocrb}} has also been installed to maximize accuracy on MRZ. The ground-truth positions of the text fields are used to obtain rectified (without perspective distortion) field images. The quadrilaterals are expanded by 2\% in each direction, relatively to the input quadrilateral, to pad the regions and reduce OCR errors.

\begin{table}[ht]
    \begin{subtable}{1.0\textwidth}
        \begin{centering}
            \begin{tabular}{|c|c|c|c|}
                \hline
                \textbf{Field type}   & \textbf{\# fields} & $\textbf{acc}_{rec}$ & \textbf{avg. CER} \\ \hline
                \textbf{Numeric}      & 6\,000              & 54.8\%               & 44.9\%            \\ \hline
                \textbf{Alpha}        & 12\,463             & 60.1\%               & 30.8\%            \\ \hline
                \textbf{Alphanumeric} & 5\,200              & 42.8\%               & 121 \%            \\ \hline
                \textbf{MRZ}          & 1\,800              & 48.6\%               & 2.52\%            \\ \hline
                \textbf{TOTAL}        & 25\,463             & 54.5\%               & 50.6\%            \\ \hline
            \end{tabular}
            \caption{per field type}
            \label{tab:tesseract_baseline_per_field_type}
        \end{centering}
    \end{subtable}
    \begin{subtable}{1.0\textwidth}
        \begin{centering}
            \begin{tabular}{|c|c|c|c|c|c|}
                \hline
                \textbf{Template}                            & \textbf{Side} & \textbf{\# fields} & $\textbf{acc}_{rec}$ & \textbf{avg. CER} \\ \hline
                \multirow{2}{*}{ \texttt{ID\_CARD\_TD1\_A} } & front         & 1\,631              & 73.8\%               & 17.0\%            \\ \cline{2-5}
                                                             & back          & 1\,000              & 81.8\%               & 1.64\%            \\ \hline
                \multirow{2}{*}{ \texttt{ID\_CARD\_TD1\_B} } & front         & 1\,832              & 61.7\%               & 59.1\%            \\ \cline{2-5}
                                                             & back          & 800                & 51.1\%               & 66.3\%            \\ \hline
                \multirow{2}{*}{ \texttt{RP\_CARD\_TD1} }    & front         & 1\,800              & 70.2\%               & 9.45\%            \\ \cline{2-5}
                                                             & back          & 1\,000              & 70.7\%               & 2.28\%            \\ \hline
                \multirow{2}{*}{ \texttt{ID\_CARD\_TD2\_A} } & front         & 1\,800              & 27.0\%               & 51.7\%            \\ \cline{2-5}
                                                             & back          & 800                & 47.5\%               & 9.36\%            \\ \hline
                \multirow{2}{*}{ \texttt{ID\_CARD\_TD2\_B} } & front         & 1\,800              & 31.3\%               & 57.9\%            \\ \cline{2-5}
                                                             & back          & 600                & 0.0\%                & 65.3\%            \\ \hline
                \multirow{2}{*}{ \texttt{RP\_CARD\_TD2} }    & front         & 2\,000              & 76.1\%               & 16.1\%            \\ \cline{2-5}
                                                             & back          & 800                & 72.5\%               & 3.43\%            \\ \hline
                \multicolumn{2}{|c|}{ \texttt{PP\_TD3\_A } } & 3\,400         & 74.9\%             & 154\%                                 \\ \hline
                \multicolumn{2}{|c|}{ \texttt{PP\_TD3\_B } } & 3\,000         & 65.0\%             & 12.2\%                                \\ \hline
                \multicolumn{2}{|c|}{ \texttt{PP\_TD3\_C } } & 3\,200         & 10.0\%             & 76.0\%                                \\ \hline
                \multicolumn{2}{|c|}{ \textbf{TOTAL} }       & 25\,463        & 54.5\%             & 50.6\%                                \\ \hline
            \end{tabular}
            \caption{per template}
            \label{tab:tesseract_baseline_per_template}
        \end{centering}
    \end{subtable}

    \caption{Text fields recognition metrics on test set, using Tesseract v5.3.3.}
    \label{tab:tesseract_baseline}
\end{table}

Table~\ref{tab:tesseract_baseline} presents the performance of this baseline. We report an $\textrm{acc}_{\textrm{rec}} = 54.5\%$, which is clearly insufficient for industrial purpose. On the MRZ specifically, Tesseract shows rather decent performance at recognizing the characters individually, with an average CER of about 2.5\%, but it is still far from being usable in an industrial context regarding the field accuracy (less that half MRZ are read perfectly). It may probably be explained by the field length (between 72 and 90 characters) and a distribution of the errors that is too uniform.

\begin{figure}[ht]
    \centering
    \includegraphics[height=0.18\textwidth]{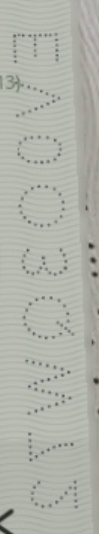}
    \includegraphics[height=0.18\textwidth]{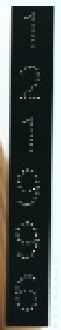}
    \includegraphics[height=0.18\textwidth]{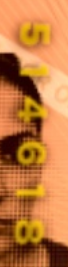}
    \begin{subfigure}[b]{0.35\textwidth}
        \centering
        \includegraphics[width=1.0\textwidth]{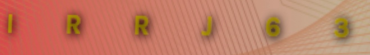} \\
        \includegraphics[width=1.0\textwidth]{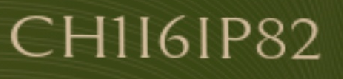} \\
    \end{subfigure}
    \begin{subfigure}[b]{0.22\textwidth}
        \centering
        \includegraphics[width=1.0\textwidth]{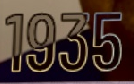} \\
    \end{subfigure}
    \caption{Some text fields that cannot be read correctly by Tesseract v5.3.3}
    \label{fig:challenging_text}
\end{figure}

Fine-tuning a dedicated OCR model on our dataset should largely improve those results. In the details, we have noticed that the most challenging fields to recognize have specific features, such as uncommon font style, rotated text, vertical text or text incrustated in the photography. Some examples of such fields are shown in Figure~\ref{fig:challenging_text}.

\section{Conclusion}

Our study introduces the DocXPand-25k dataset, which we think is promising for the development and evaluation of document analysis methods, contributing to the ongoing advancement of automatic identity verification technologies. It encompasses a diverse range of document types and backgrounds, and aims to closely mimic real-world distributions of ID images. This resemblance is validated by the LPIPS-based metric, which shows that our proposed dataset closely aligns with end-users made captures (photos, scans and screenshots) of real IDs in terms of visual similarity. While we provide baselines for ID classification, ID localization and text field recognition, we recognize the scope for further enhancements and research. We particularly encourage researchers to use and adapt the tools we developed to construct this dataset to make the benchmarks more challenging.

\section*{Acknowledgments}

We extend our heartfelt gratitude to Alexis Bergès and Frédéric Guay for their proposal, organization, and management of this project, all of which laid the foundation for this research.
Our appreciation also goes to QuickSign for granting us access to computational resources, which proved instrumental for our experiments.

\printbibliography

@article{arlazarov2019midv,
  title={MIDV-500: a dataset for identity document analysis and recognition on mobile devices in video stream},
  author={Arlazarov, Vladimir and Bulatov, Konstantin and Chernov, Timofey and Arlazarov, Vladimir},
  journal={Computer Optics},
  volume={43},
  number={5},
  pages={818--824},
  year={2019}
}

@inproceedings{bulatov2020midv,
  title={MIDV-2019: challenges of the modern mobile-based document OCR},
  author={Bulatov, Konstantin and Matalov, Daniil and Arlazarov, Vladimir},
  booktitle={Twelfth International Conference on Machine Vision (ICMV 2019)},
  volume={11433},
  pages={717--722},
  year={2020},
  organization={SPIE}
}

@article{bulatov2022midv,
  title={MIDV-2020: A comprehensive benchmark dataset for identity document analysis},
  author={Bulatov, Konstantin and Emelianova, Ekaterina and Tropin, Daniil and Skoryukina, Natalya and Chernyshova, Yulia and Ming, Zuheng and Burie, Jean-Christophe and Luqman, Muhammad},
  journal={Computer Optics},
  volume={46},
  number={2},
  pages={252--270},
  year={2022}
}

@inproceedings{chiron2021id,
  title={ID documents matching and localization with multi-hypothesis constraints},
  author={Chiron, Guillaume and Ghanmi, Nabil and Awal, Ahmad Montaser},
  booktitle={2020 25th International Conference on Pattern Recognition (ICPR)},
  pages={3644--3651},
  year={2021},
  organization={IEEE}
}

@article{polevoy2022document,
  title={Document Liveness Challenge Dataset (DLC-2021)},
  author={Polevoy, Dmitry V and Sigareva, Irina V and Ershova, Daria M and Arlazarov, Vladimir V and Nikolaev, Dmitry P and Ming, Zuheng and Luqman, Muhammad Muzzamil and Burie, Jean-Christophe},
  journal={Journal of Imaging},
  volume={8},
  number={7},
  pages={181},
  year={2022},
  publisher={MDPI}
}

@inproceedings{attivissimo2019automatic,
  title={An automatic reader of identity documents},
  author={Attivissimo, Filippo and Giaquinto, Nicola and Scarpetta, Marco and Spadavecchia, Maurizio},
  booktitle={2019 IEEE International Conference on Systems, Man and Cybernetics (SMC)},
  pages={3525--3530},
  year={2019},
  organization={IEEE}
}

@inproceedings{chernyshova2021midv,
  title={MIDV-LAIT: a challenging dataset for recognition of IDs with Perso-Arabic, Thai, and Indian scripts},
  author={Chernyshova, Yulia and Emelianova, Ekaterina and Sheshkus, Alexander and Arlazarov, Vladimir V},
  booktitle={International Conference on Document Analysis and Recognition},
  pages={258--272},
  year={2021},
  organization={Springer}
}

@inproceedings{skoryukina2019fast,
  title={Fast method of ID documents location and type identification for mobile and server application},
  author={Skoryukina, Natalya and Arlazarov, Vladimir and Nikolaev, Dmitry},
  booktitle={2019 International Conference on Document Analysis and Recognition (ICDAR)},
  pages={850--857},
  year={2019},
  organization={IEEE}
}

@inproceedings{bulatov2017smart,
  title={Smart IDReader: Document recognition in video stream},
  author={Bulatov, Konstantin and Arlazarov, Vladimir V and Chernov, Timofey and Slavin, Oleg and Nikolaev, Dmitry},
  booktitle={2017 14th IAPR International Conference on Document Analysis and Recognition (ICDAR)},
  volume={6},
  pages={39--44},
  year={2017},
  organization={IEEE}
}

@article{wu2022ldrnet,
  title={LDRNet: Enabling Real-time Document Localization on Mobile Devices},
  author={Wu, Han and Qian, Holland and Wu, Huaming},
  journal={arXiv preprint arXiv:2206.02136},
  year={2022}
}

@misc{prado2022,
    title={Public Register of Authentic identity and travel Documents Online},
    url={https://www.consilium.europa.eu/prado/en/prado-start-page.html},
    publisher={Council of the European Union},
    year={2023},
    month={08}
}

@techreport{doc9303,
    type = {Specifications},
    key = {ICAO Doc 9303},
    year = {2021},
    title = {Machine Readable Travel Documents},
    number = {Eighth edition},
    institution = {International Civil Aviation Organization}
}

@inproceedings{ngoc2018saliency,
  title={Saliency-based detection of identy documents captured by smartphones},
  author={Ngoc, Minh {\^O}n V{\~u} and Fabrizio, Jonathan and G{\'e}raud, Thierry},
  booktitle={2018 13th IAPR International Workshop on Document Analysis Systems (DAS)},
  pages={387--392},
  year={2018},
  organization={IEEE}
}

@inproceedings{de2020bid,
  title={BID Dataset: a challenge dataset for document processing tasks},
  author={de S{\'a} Soares, Alysson and das Neves Junior, Ricardo Batista and Bezerra, Byron Leite Dantas},
  booktitle={Anais Estendidos do XXXIII Conference on Graphics, Patterns and Images},
  pages={143--146},
  year={2020},
  organization={SBC}
}

@article{reinhard2001color,
  title={Color transfer between images},
  author={Reinhard, Erik and Adhikhmin, Michael and Gooch, Bruce and Shirley, Peter},
  journal={IEEE Computer graphics and applications},
  volume={21},
  number={5},
  pages={34--41},
  year={2001},
  publisher={IEEE}
}

@inproceedings{rombach2022high,
  title={High-resolution image synthesis with latent diffusion models},
  author={Rombach, Robin and Blattmann, Andreas and Lorenz, Dominik and Esser, Patrick and Ommer, Bj{\"o}rn},
  booktitle={Proceedings of the IEEE/CVF conference on computer vision and pattern recognition},
  pages={10684--10695},
  year={2022}
}

@inproceedings{smith2007overview,
  title={An overview of the Tesseract OCR engine},
  author={Smith, Ray},
  booktitle={Ninth international conference on document analysis and recognition (ICDAR 2007)},
  volume={2},
  pages={629--633},
  year={2007},
  organization={IEEE}
}

@article{kabeshova2024sdlnet,
  author    = {Anastasiia Kabeshova and Guillaume Betmont and Julien Lerouge and Evgeny Stepankevich and Alexis Bergès},
  title     = {SDL-Net pour la Localisation de Documents Structurés : un U-Net économe en ressources.},
  journal = {Revue des Nouvelles Technologies de l'Information},
  volume = {Extraction et Gestion des Connaissances, RNTI-E-40},
  year      = {2024},
  pages     = {295-302}
}

@article{tropin2021advanced,
  title={Advanced Hough-based method for on-device document localization},
  author={Tropin, Vyacheslavovich Daniil and Ershov Alexandr Mikhailovich and Nikolaev, Dmitry Petrovich and Arlazarov, Vladimir Viktorovich},
  journal={Computer optics},
  volume={45},
  number={5},
  pages={702--712},
  year={2021}
}

@article{zhu2019coarse,
  title={Coarse-to-fine document localization in natural scene image with regional attention and recursive corner refinement},
  author={Zhu, Anna and Zhang, Chen and Li, Zhi and Xiong, Shengwu},
  journal={International Journal on Document Analysis and Recognition (IJDAR)},
  volume={22},
  pages={351--360},
  year={2019},
  publisher={Springer}
}

@inproceedings{ngoc2019document,
  title={Document detection in videos captured by smartphones using a saliency-based method},
  author={Ngoc, Minh {\^O}n V{\~u} and Fabrizio, Jonathan and G{\'e}raud, Thierry},
  booktitle={2019 International Conference on Document Analysis and Recognition Workshops (ICDARW)},
  volume={4},
  pages={19--24},
  year={2019},
  organization={IEEE}
}

@inproceedings{zhang2018unreasonable,
  title={The unreasonable effectiveness of deep features as a perceptual metric},
  author={Zhang, Richard and Isola, Phillip and Efros, Alexei A and Shechtman, Eli and Wang, Oliver},
  booktitle={Proceedings of the IEEE conference on computer vision and pattern recognition},
  pages={586--595},
  year={2018}
}

@inproceedings{deng2009imagenet,
  title={Imagenet: A large-scale hierarchical image database},
  author={Deng, Jia and Dong, Wei and Socher, Richard and Li, Li-Jia and Li, Kai and Fei-Fei, Li},
  booktitle={2009 IEEE conference on computer vision and pattern recognition},
  pages={248--255},
  year={2009},
  organization={Ieee}
}

@inproceedings{tan2021efficientnetv2,
  title={Efficientnetv2: Smaller models and faster training},
  author={Tan, Mingxing and Le, Quoc},
  booktitle={International conference on machine learning},
  pages={10096--10106},
  year={2021},
  organization={PMLR}
}

@inproceedings{howard2019searching,
  title={Searching for mobilenetv3},
  author={Howard, Andrew and Sandler, Mark and Chu, Grace and Chen, Liang-Chieh and Chen, Bo and Tan, Mingxing and Wang, Weijun and Zhu, Yukun and Pang, Ruoming and Vasudevan, Vijay and others},
  booktitle={Proceedings of the IEEE/CVF international conference on computer vision},
  pages={1314--1324},
  year={2019}
}

\newpage

\appendix

\section{Supplementary Material}

\subsection{Specifications of our fictitious MROTD classes}
\label{sec:specifications}

For each fictitious MROTD class we had to design, we took inspiration from real documents issued by European Union states, and created a list of text fields (such as names, places, dates and various identifiers), picture contents (identity photo, ghost image, barcodes), visual features (such as OVI, plastic border and background patterns) that the class should contain, to maximize the realism of the designed templates. Our other goal was also to guarantee the diverseness of the templates, therefore we chose many document sizes and general disposition of the content, i.e. the TD1, TD2 and TD3 types, and tried to represent most of the difficult cases we had to handle internally when automatically reading real identity documents, such as :
\begin{itemize}
    \item merged first name and last name fields without a textual separator
    \item multiple templates of Belgian documents (one for each of the three official languages)
    \item textual patterns overlaid over the text fields ("RF" band on old French ID cards)
\end{itemize}

Table~\ref{tab:specifications} is a small excerpt of the specifications we gave to our graphic designer to make the vectorial templates.

\begin{table}[ht]
    \begin{adjustbox}{width=\textwidth}
        \begin{tabular}{|c|c|c|c|c|c|c|}
            \hline
            \textbf{Template}                            & \textbf{Type}                      & \textbf{Format} & \textbf{Main language}      & \textbf{Additional translations} & \textbf{Special content}  & \textbf{Special layout or visual features} \\ \hline
            \multirow{2}{*}{ \texttt{ID\_CARD\_TD1\_A} } & Identity card                      & TD1             & German                      & English, French                  & Data matrix, ghost image, & Complex background, many overlaid          \\
                                                         & (biometric, with electronic chip)  &                 &                             &                                  & long document number      & textures and OVI                           \\ \hline
            \multirow{2}{*}{ \texttt{ID\_CARD\_TD1\_B} } & Identity card                      & TD1             & English                     & \textit{NA}                      & Ghost image               & Vertical, overspaced,                      \\
                                                         & (biometric, with hidden RFID chip) &                 &                             &                                  &                           & low-contrast text fields                   \\ \hline
            \multirow{2}{*}{ \texttt{RP\_CARD\_TD1} }    & Residence permit                   & TD1             & Dutch or French or German   & English                          & Merged last name and      & Variable layout language,                  \\
                                                         & (biometric, with electronic chip)  &                 & (depending on the document) &                                  & first name fields         & complex overlaid textures and OVI          \\ \hline
            \multirow{2}{*}{ \texttt{ID\_CARD\_TD2\_A} } & Identity card                      & TD2             & French                      & \textit{NA}                      & Long document number      & Transparent borders,                       \\
                                                         &                                    &                 &                             &                                  &                           & band of "RP" pattern                       \\ \hline
            \multirow{2}{*}{ \texttt{ID\_CARD\_TD2\_B} } & Identity card                      & TD2             & Spanish                     & Catalan                          & Barcode, ghost image      & Transparent borders,                       \\
                                                         &                                    &                 &                             &                                  &                           & little space between text fields           \\ \hline
            \multirow{2}{*}{ \texttt{RP\_CARD\_TD2} }    & Residence permit                   & TD2             & Italian                     & English                          & Merged last name and      & \textit{NA}                                \\
                                                         &                                    &                 &                             &                                  & First name fields         &                                            \\ \hline
            \multirow{2}{*}{ \texttt{PP\_TD3\_A} }       & Passport                           & TD3             & French                      & English                          & Ghost image               & \textit{NA}                                \\
                                                         &                                    &                 &                             &                                  &                           &                                            \\ \hline
            \multirow{2}{*}{ \texttt{PP\_TD3\_B} }       & Passport                           & TD3             & Portuguese                  & French                           & \textit{NA}               & \textit{NA}                                \\
                                                         &                                    &                 &                             &                                  &                           &                                            \\ \hline
            \multirow{2}{*}{ \texttt{PP\_TD3\_C} }       & Passport                           & TD3             & Dutch                       & French                           & Ghost image,              & \textit{NA}                                \\
                                                         &                                    &                 &                             &                                  & date in photo             &                                            \\ \hline
        \end{tabular}
    \end{adjustbox}
    \normalsize
    \caption{Brief specifications used to design the vectorial templates.}
    \label{tab:specifications}
\end{table}

Table~\ref{tab:field_categories} present the detailed list of text fields included on each side of the document classes, shows their diversity of text ifle  categories of text fields used for each ID class, and .

\begin{table}[ht]
    \begin{adjustbox}{width=\textwidth}
        \begin{tabular}{|c|c|c|c|c|c|}
            \hline
            \textbf{Template}                            & \textbf{Side}                        & \textbf{Numeric}              & \textbf{Alpha}                                        & \textbf{Alphanumeric}                     & \textbf{MRZ}  \\ \hline
            \multirow{2}{*}{ \texttt{ID\_CARD\_TD1\_A} } & front                                & \makecell{birth\_date,                                                                                                                            \\ card\_access\_number, expires }                                    & \makecell{birth\_name, birth\_place, family\_name, \\ given\_name, nationality}                               & document\_number                                                                  & \textit{NA}   \\ \cline{2-6}
                                                         & back                                 & date\_issued                  & authority                                             & address, height                           & MRZ (3 lines) \\ \hline
            \multirow{2}{*}{ \texttt{ID\_CARD\_TD1\_B} } & front                                & card\_access\_number, expires & \makecell{birth\_name, family\_name, gender,                                                                      \\ given\_name, nationality}                                     & \makecell{birth\_date\_and\_place, \\ document\_number, \\ personal\_number}                       & \textit{NA}   \\ \cline{2-6}
                                                         & back                                 & card\_access\_number          & authority                                             & date\_and\_place\_issued                  & MRZ (3 lines) \\ \hline
            \multirow{2}{*}{ \texttt{RP\_CARD\_TD1} }    & front                                & birth\_date, expires          & \makecell{family\_name, foreign\_nationality, gender,                                                             \\ given\_name, observations, permit\_type}             & document\_number                                                                  & \textit{NA}   \\ \cline{2-6}
                                                         & back                                 & date\_issued                  & birth\_place, observations, place\_issued             & \textit{NA}                               & MRZ (3 lines) \\ \hline
            \multirow{2}{*}{ \texttt{ID\_CARD\_TD2\_A} } & front                                & birth\_date                   & family\_name, gender, given\_name, initials           & \makecell{birth\_place, document\_number,                 \\ height}                                            & MRZ (2 lines) \\ \cline{2-6}
                                                         & back                                 & date\_issued, expires         & \textit{NA}                                           & address, authority                        & \textit{NA}   \\ \hline
            \multirow{2}{*}{ \texttt{ID\_CARD\_TD2\_B} } & front                                & birth\_date, expires          & \makecell{family\_name, gender, given\_name,                                                                      \\ nationality, second\_family\_name}                            & document\_number                                                                  & MRZ (2 lines) \\ \cline{2-6}
                                                         & back                                 & \textit{NA}                   & birth\_place, parents                                 & address                                   & \textit{NA}   \\ \hline
            \multirow{2}{*}{ \texttt{RP\_CARD\_TD2} }    & front                                & birth\_date, expires          & \makecell{family\_name, foreign\_nationality, gender,                                                             \\ given\_name, observations, permit\_type}             & document\_number                                                                  & MRZ (2 lines) \\ \cline{2-6}
                                                         & back                                 & date\_issued                  & birth\_place, observations, place\_issued             & \textit{NA}                               & \textit{NA}   \\ \hline
            \multicolumn{2}{|c|}{ \texttt{PP\_TD3\_A } } & \makecell{birth\_date, date\_issued,                                                                                                                                                     \\ expires}                                             & \makecell{country\_code, eyes\_color, family\_name, \\ gender, given\_name, nationality, type}                & \makecell{address, authority, \\ birth\_place, document\_number, \\ document\_number\_bis, height} &               \\ \cline{1-5}
            \multicolumn{2}{|c|}{ \texttt{PP\_TD3\_B } } & \makecell{birth\_date, date\_issued,                                                                                                                                                     \\ expires, personal\_number }                          & \makecell{authority, birth\_place, country\_code, \\ family\_name, gender, given\_name, \\ nationality, type} & document\_number, height                                                          & MRZ (2 lines) \\ \cline{1-5}
            \multicolumn{2}{|c|}{ \texttt{PP\_TD3\_C } } & \makecell{birth\_date, date\_issued,                                                                                                                                                     \\ expires, photo\_birth\_month, \\ photo\_birth\_year} & \makecell{authority, birth\_place, country\_code, \\ family\_name, gender, given\_name, \\ nationality, type} & document\_number, height                                                          &               \\ \hline
        \end{tabular}
    \end{adjustbox}
    \normalsize
    \caption{Detailed specifications of text fields included on each side.}
    \label{tab:field_categories}
\end{table}

\subsection{Generated document samples}
\label{sec:docxpand_samples}

\begin{figure}[ht]\captionsetup[subfigure]{font=scriptsize}
    \centering
    \begin{subfigure}[b]{0.32\textwidth}
        \centering
        \includegraphics[width=\textwidth]{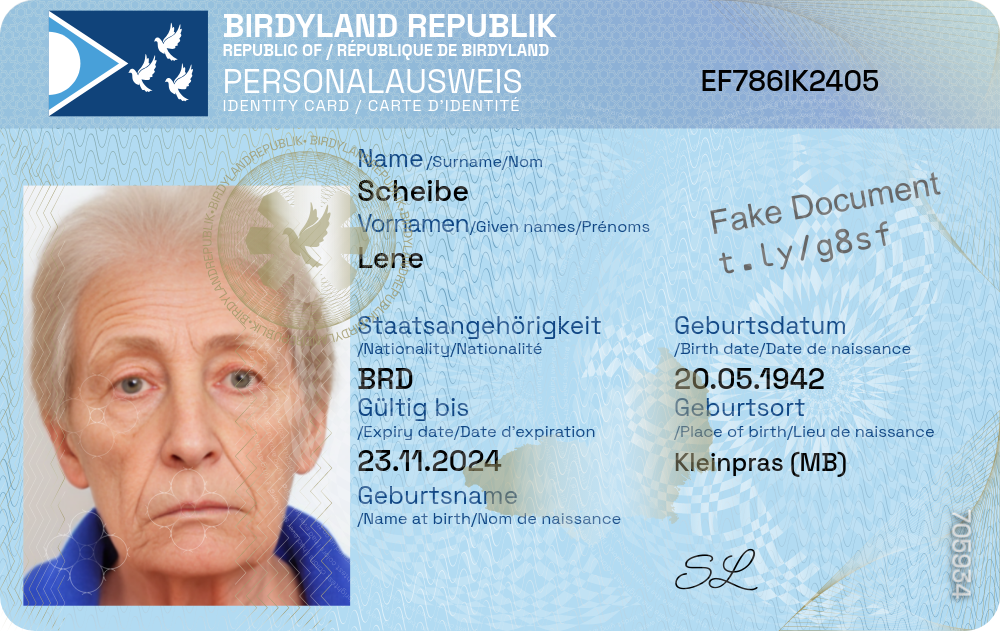}
        \caption*{ID\_CARD\_TD1\_A (front)}
    \end{subfigure}
    \hfill
    \begin{subfigure}[b]{0.32\textwidth}
        \centering
        \includegraphics[width=\textwidth]{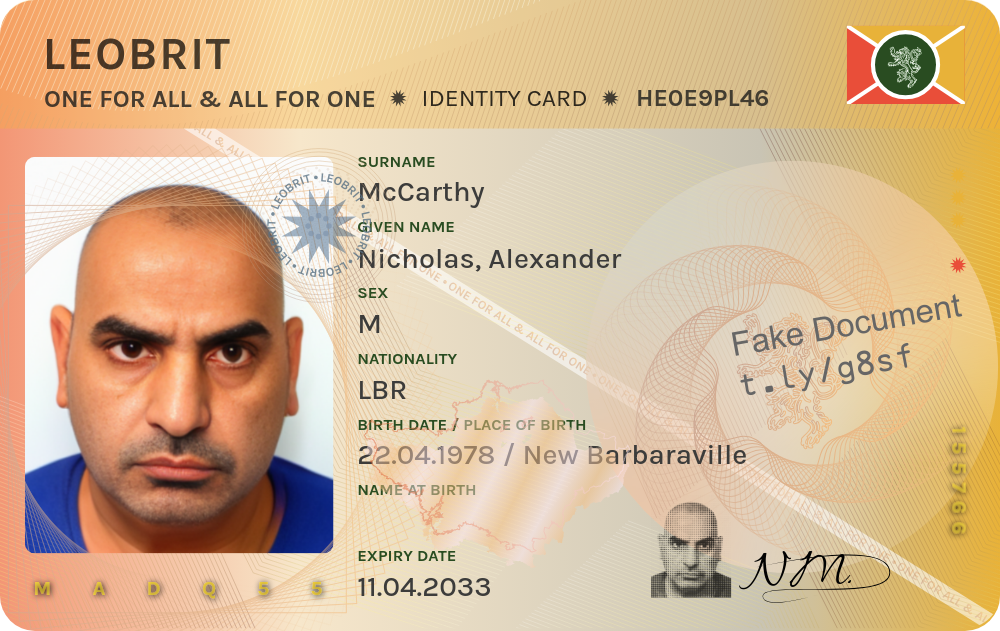}
        \caption*{ID\_CARD\_TD1\_B (front)}
    \end{subfigure}
    \hfill
    \begin{subfigure}[b]{0.32\textwidth}
        \centering
        \includegraphics[width=\textwidth]{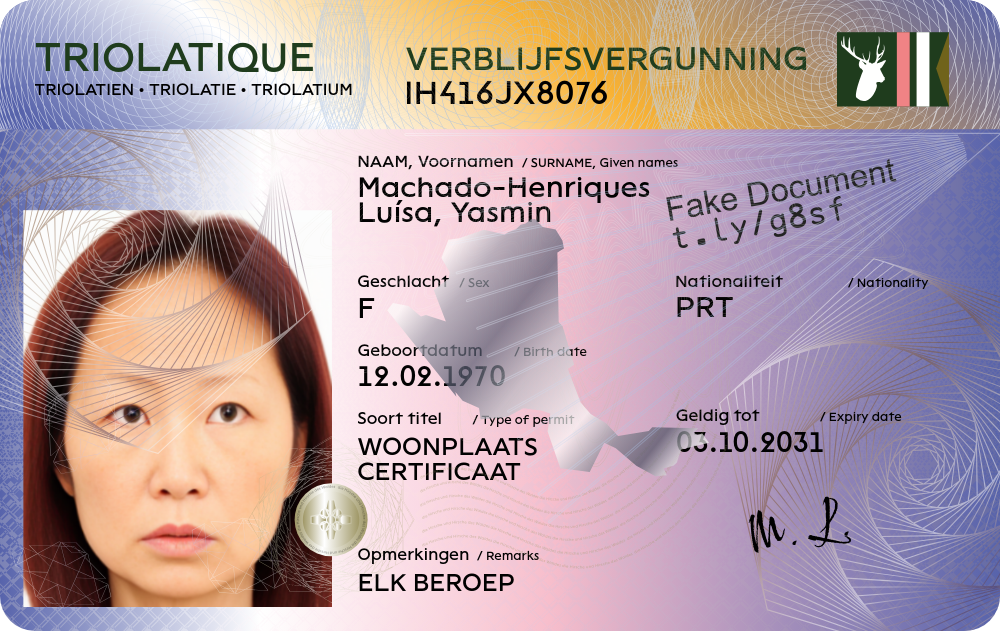}
        \caption*{RP\_CARD\_TD1 (front)}
    \end{subfigure}
    \\
    \begin{subfigure}[b]{0.32\textwidth}
        \centering
        \includegraphics[width=\textwidth]{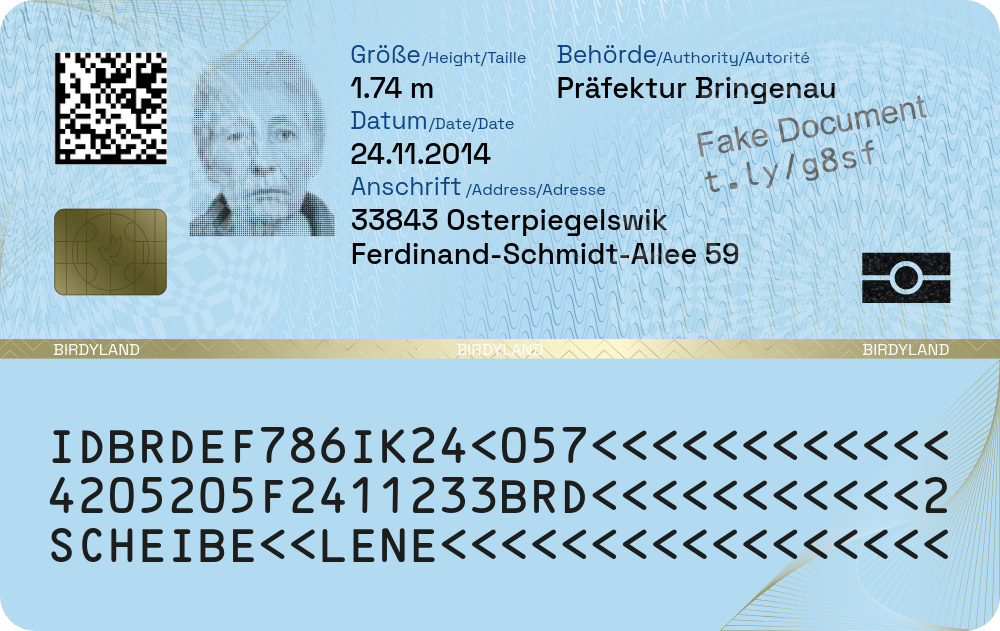}
        \caption*{ID\_CARD\_TD1\_A (back)}
    \end{subfigure}
    \hfill
    \begin{subfigure}[b]{0.32\textwidth}
        \centering
        \includegraphics[width=\textwidth]{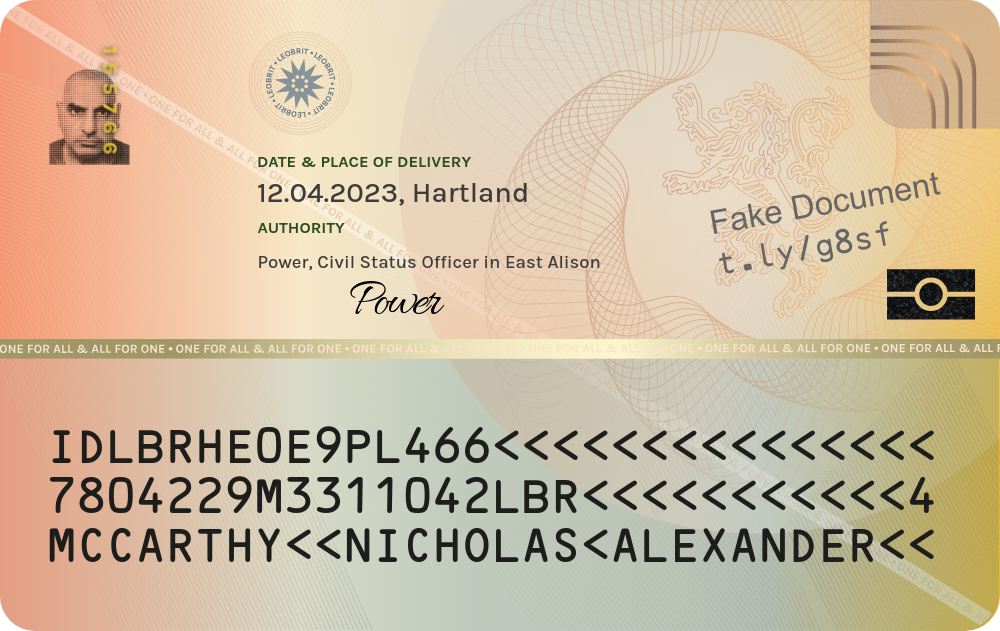}
        \caption*{ID\_CARD\_TD1\_B (back)}
    \end{subfigure}
    \hfill
    \begin{subfigure}[b]{0.32\textwidth}
        \centering
        \includegraphics[width=\textwidth]{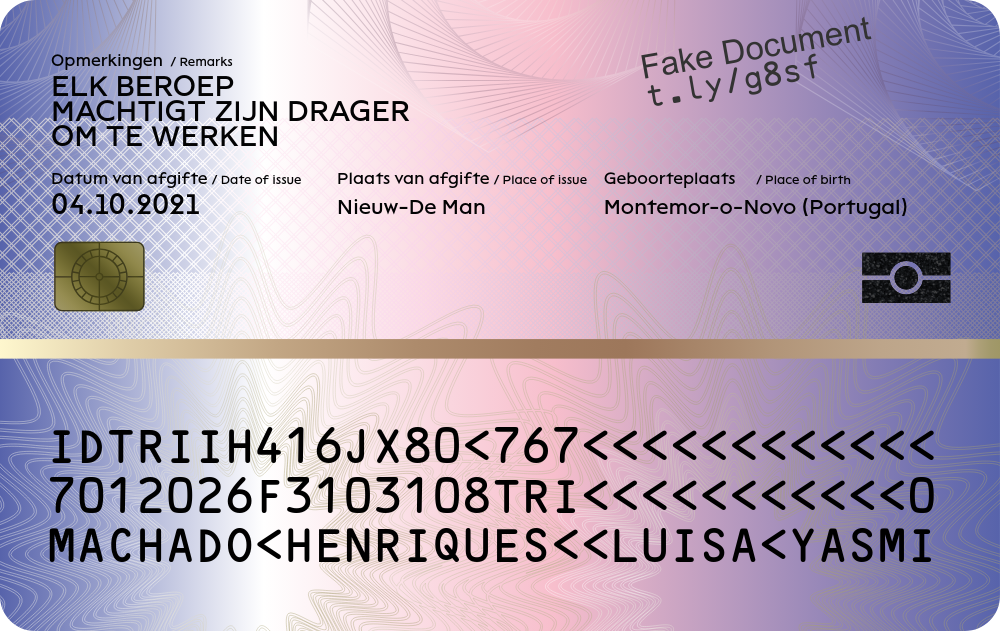}
        \caption*{RP\_CARD\_TD1 (back)}
    \end{subfigure}
    \\
    \begin{subfigure}[b]{0.32\textwidth}
        \centering
        \includegraphics[width=\textwidth]{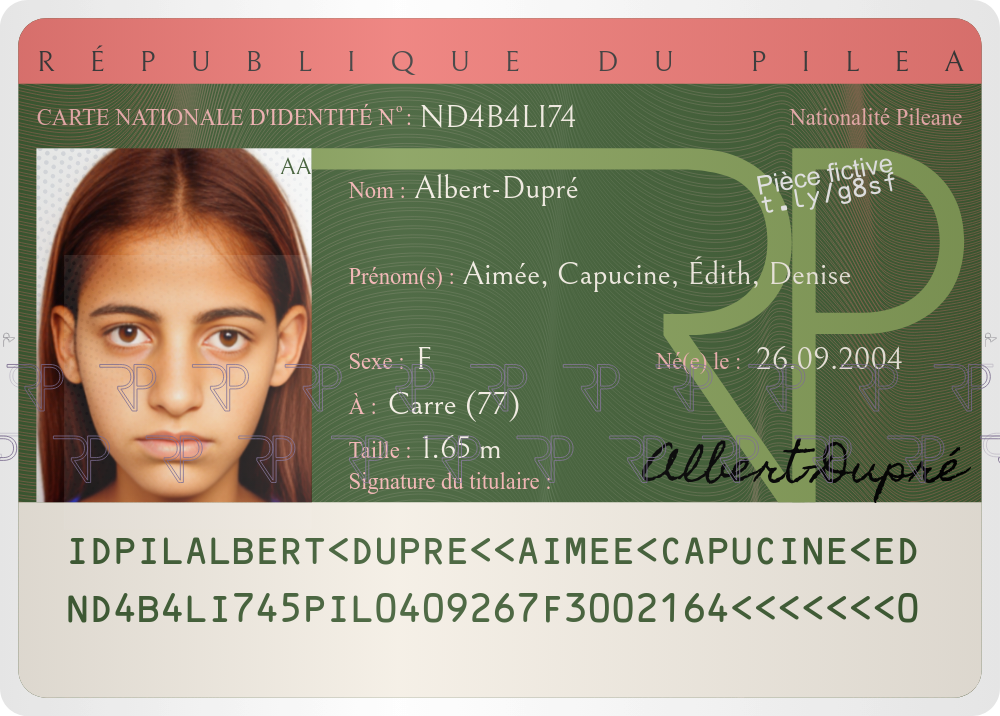}
        \caption*{ID\_CARD\_TD2\_A (front)}
    \end{subfigure}
    \hfill
    \begin{subfigure}[b]{0.32\textwidth}
        \centering
        \includegraphics[width=\textwidth]{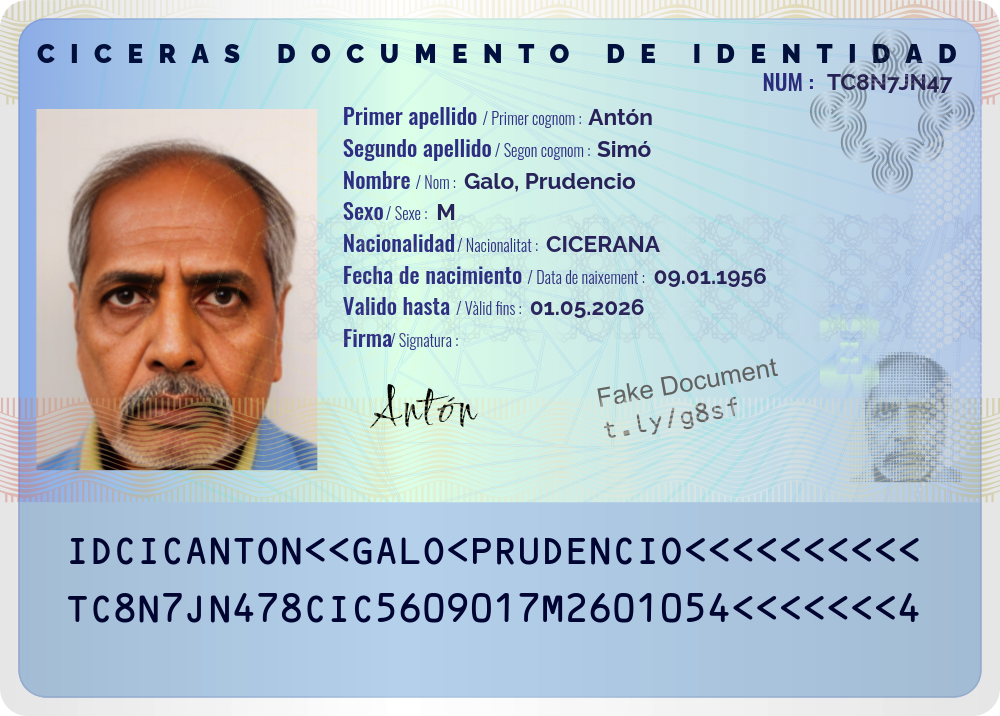}
        \caption*{ID\_CARD\_TD2\_B (front)}
    \end{subfigure}
    \hfill
    \begin{subfigure}[b]{0.32\textwidth}
        \centering
        \includegraphics[width=\textwidth]{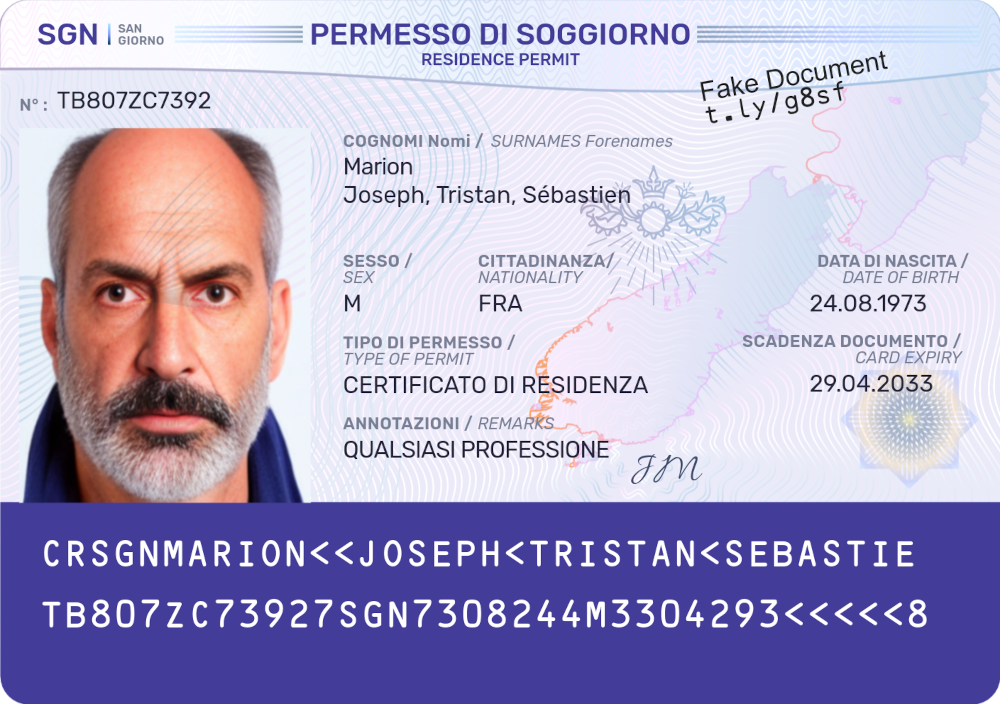}
        \caption*{RP\_CARD\_TD2 (front)}
    \end{subfigure}
    \\
    \begin{subfigure}[b]{0.32\textwidth}
        \centering
        \includegraphics[width=\textwidth]{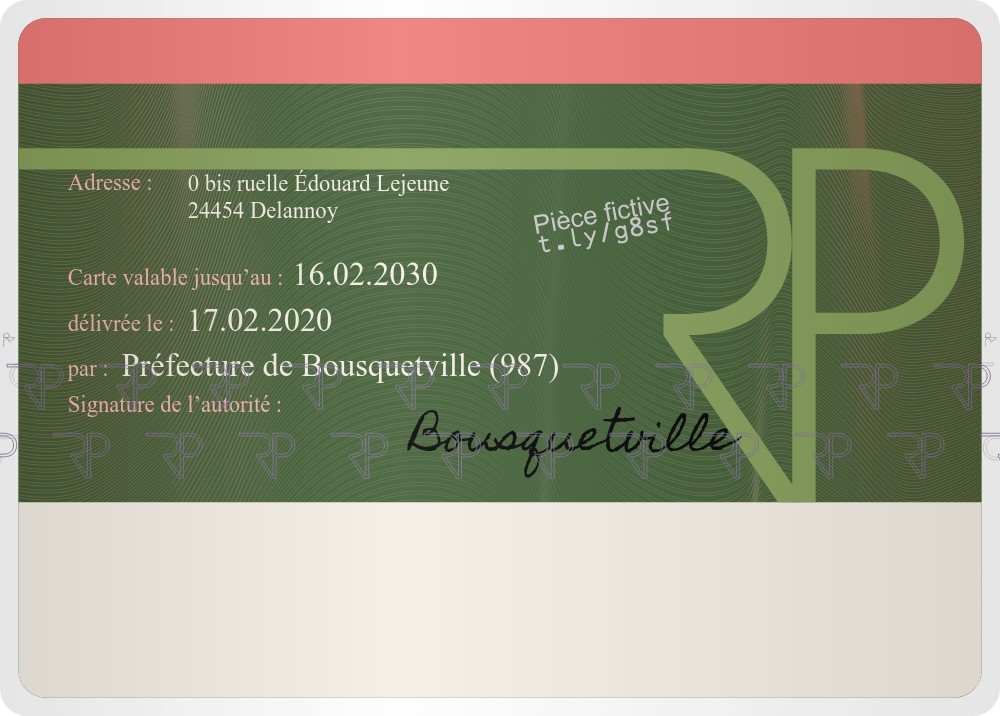}
        \caption*{ID\_CARD\_TD2\_A (back)}
    \end{subfigure}
    \hfill
    \begin{subfigure}[b]{0.32\textwidth}
        \centering
        \includegraphics[width=\textwidth]{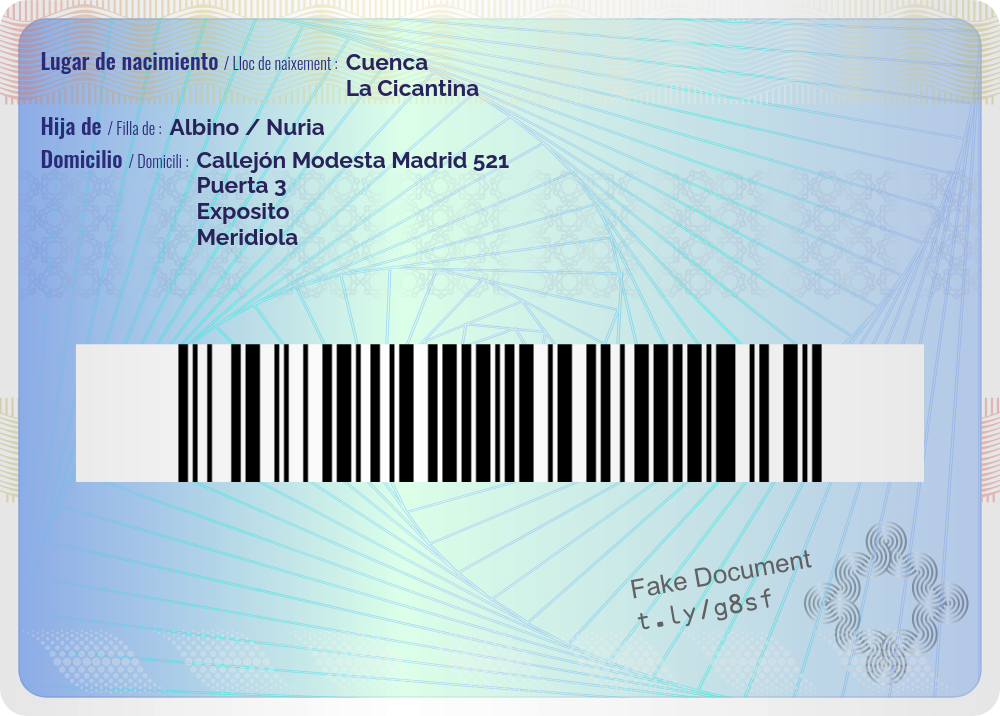}
        \caption*{ID\_CARD\_TD2\_B (back)}
    \end{subfigure}
    \hfill
    \begin{subfigure}[b]{0.32\textwidth}
        \centering
        \includegraphics[width=\textwidth]{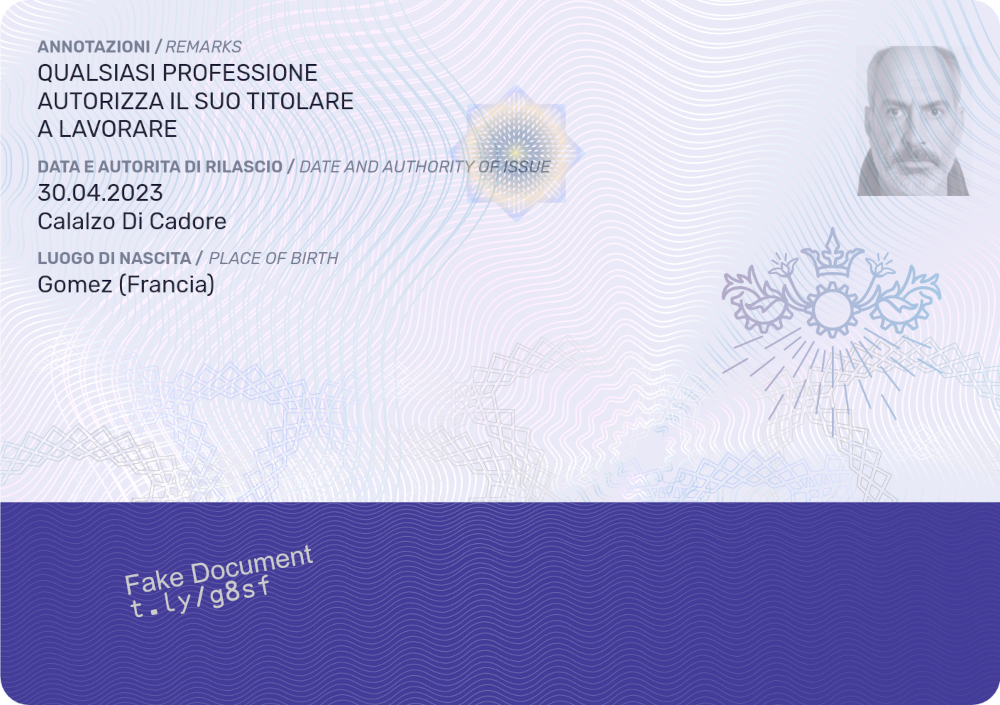}
        \caption*{RP\_CARD\_TD2 (back)}
    \end{subfigure}
    \\
    \begin{subfigure}[b]{0.32\textwidth}
        \centering
        \includegraphics[width=\textwidth]{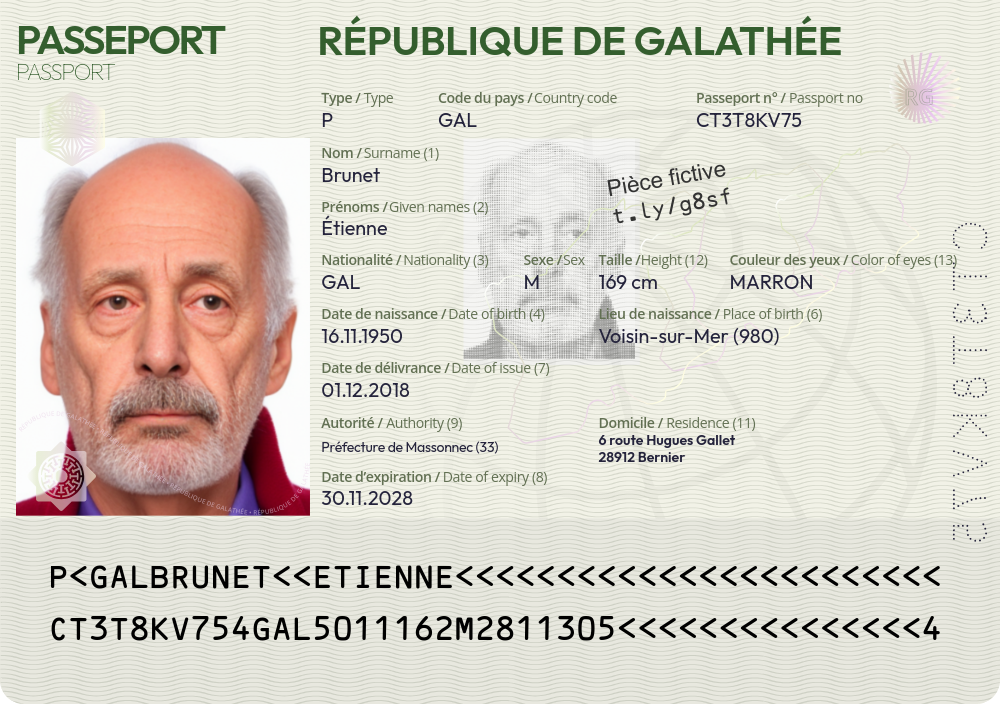}
        \caption*{PP\_TD3\_A}
    \end{subfigure}
    \hfill
    \begin{subfigure}[b]{0.32\textwidth}
        \centering
        \includegraphics[width=\textwidth]{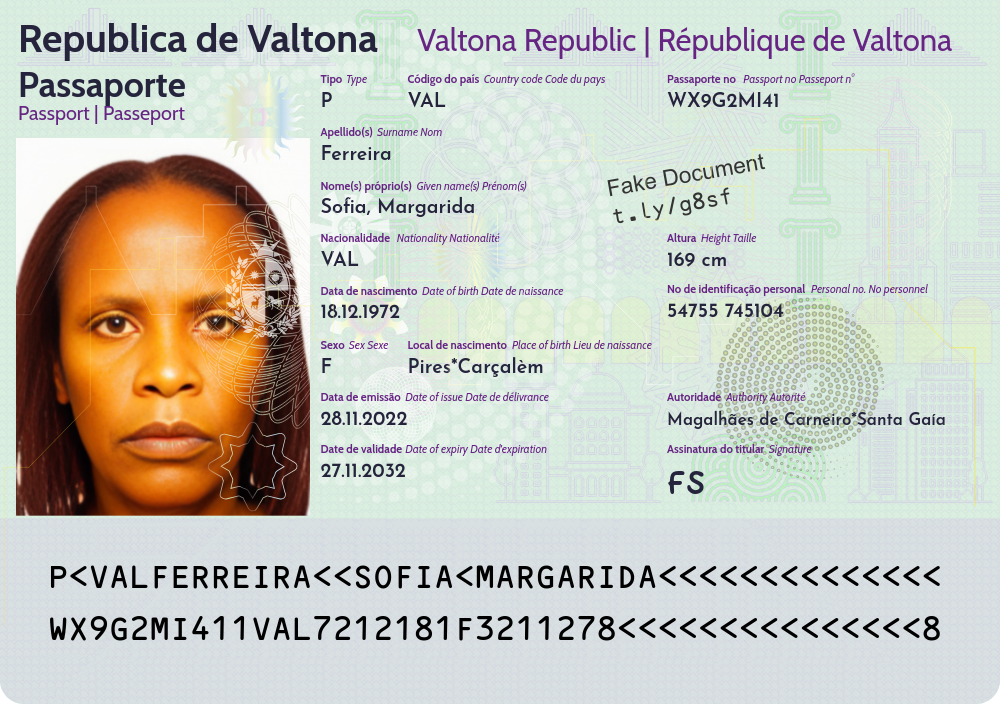}
        \caption*{PP\_TD3\_B}
    \end{subfigure}
    \hfill
    \begin{subfigure}[b]{0.32\textwidth}
        \centering
        \includegraphics[width=\textwidth]{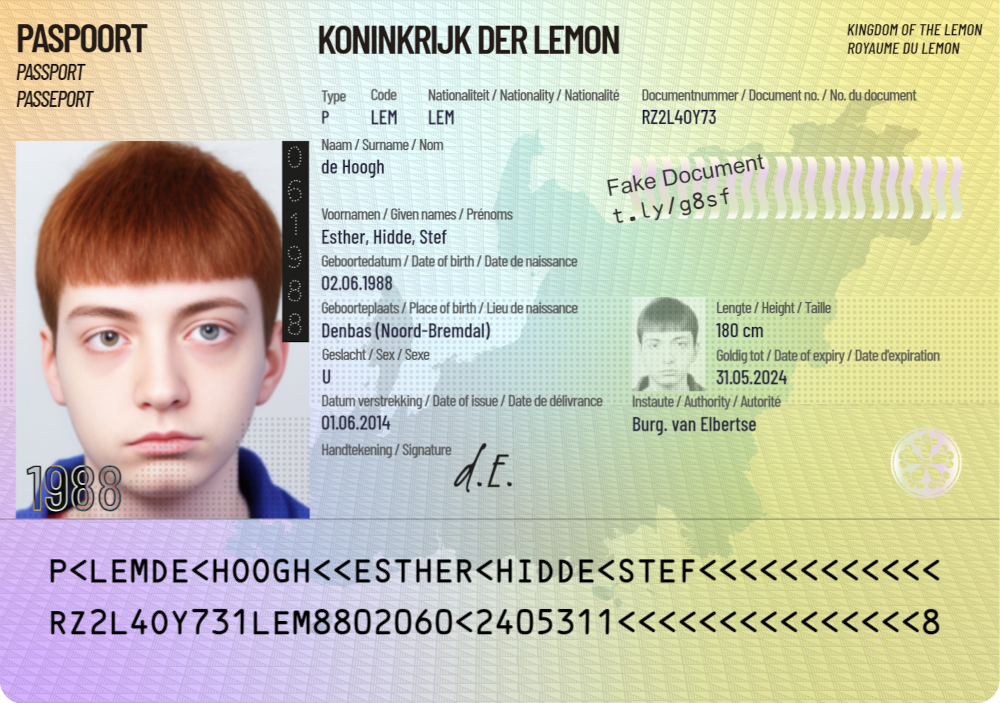}
        \caption*{PP\_TD3\_C}
    \end{subfigure}
    \normalsize
    \caption{Samples of generated documents for each class.}
    \label{fig:generated_docs_samples}
\end{figure}

Figure~\ref{fig:generated_docs_samples} shows samples of generated documents for each of DocXPand class. These documents are ready to be integrated into scenes. Some results of this integration process is illustrated in Figure~\ref{fig:integrated_docs_samples}

\subsection{Labels sample}

\begin{code}
    \captionof{listing}{Sample of labels in JSON dataset. Field positions have been abridged for conciseness.}
    \label{lst:labels}
    \begin{minted}[numbersep=5pt,
               frame=lines,
               gobble=4,
               framesep=2mm,
               fontsize=\tiny]{json}
{
   "_id":"2e39bebb-7c3f-4e30-b765-21b91ec59fda-front",
   "annotations":[
      {
         "_id":0,
         "annotator":"automatic",
         "created_at":"2023-07-13T13:11:11.603388",
         "fields":{
            "front":{
               "birth_date":{
                  "position":{ "p1": { "x": ... }, ... },
                  "type":"text",
                  "value":"06.04.1981"
               },
               "birth_place":{
                  "position":{ "p1": { "x": ... }, ... },
                  "type":"text",
                  "value":"Leclercq (70)"
               },
               "document_number":{
                  "position":{ "p1": { "x": ... }, ... },
                  "type":"text",
                  "value":"US7M1NV39"
               },
               "family_name":{
                  "position":{ "p1": { "x": ... }, ... },
                  "type":"text",
                  "value":[
                     "Juncken"
                  ]
               },
               "gender":{
                  "position":{ "p1": { "x": ... }, ... },
                  "type":"text",
                  "value":"M"
               },
               "given_name":{
                  "position":{ "p1": { "x": ... }, ... },
                  "type":"text",
                  "value":[
                     "Dierk, Filippo"
                  ]
               },
               "height":{
                  "position":{ "p1": { "x": ... }, ... },
                  "type":"text",
                  "value":"1.84 m"
               },
               "initials":{
                  "position":{ "p1": { "x": ... }, ... },
                  "type":"text",
                  "value":"JD"
               },
               "mrz":{
                  "position":{ "p1": { "x": ... }, ... },
                  "type":"text",
                  "value":[
                     "IDPILJUNCKEN<<DIERK<FILIPPO<<<<<<<<<",
                     "US7M1NV396PIL8106045M2802046<<<<<<<2"
                  ]
               },
               "nationality":{
                  "position":null,
                  "type":"text",
                  "value":"PIL"
               },
               "photo":{
                  "position":{ "p1": { "x": ... }, ... },
                  "type":"image"
                  "filename":"2e39bebb-7c3f-4e30-b765-21b91ec59fda-ID_CARD_TD2_A-front-photo.jpg",
               },
               "signature":{
                  "position":{ "p1": { "x": ... }, ... },
                  "type":"text",
                  "value":"J.D."
               }
            }
         },
         "position":{
            "p1":{
               "x":0.0846774193548387,
               "y":0.30350665054413545
            },
            "p2":{
               "x":0.9379032258064516,
               "y":0.2708585247883918
            },
            "p3":{
               "x":0.957258064516129,
               "y":0.6717049576783555
            },
            "p4":{
               "x":0.11048387096774194,
               "y":0.7176541717049577
            }
         },
         "scene_image":"ec6ecbc1d6a819c1537ee575ae8aed42",
         "template":"ID_CARD_TD2_A-front",
         "updated_at":"2023-07-25T21:34:28.654907"
      }
   ],
   "filename":"ID_CARD_TD2_A/2e39bebb-7c3f-4e30-b765-21b91ec59fda-ID_CARD_TD2_A-front.jpg"
}
\end{minted}
\end{code}

\end{document}